\documentclass{article} % For LaTeX2e
\usepackage{ICLR2026/iclr2026_conference,times}

\usepackage[utf8]{inputenc} % allow utf-8 input
\usepackage[T1]{fontenc}    % use 8-bit T1 fonts
\usepackage{hyperref}       % hyperlinks
\usepackage{url}            % simple URL typesetting
\usepackage{booktabs}       % professional-quality tables
\usepackage{amsfonts}       % blackboard math symbols
\usepackage{nicefrac}       % compact symbols for 1/2, etc.
\usepackage{microtype}      % microtypography
\usepackage{xcolor}         % colors
\usepackage{array}
\usepackage{subcaption}
\usepackage{wrapfig}
\usepackage{appendix}
\usepackage{graphicx}
\usepackage{xspace}
\usepackage{amsmath}
\usepackage{amssymb}
\usepackage{mathtools}
\usepackage{amsthm}
\usepackage[table]{xcolor}
\usepackage{algorithm}
\usepackage{algorithmic}
\usepackage[capitalize,noabbrev]{cleveref}
\Crefname{ALC@unique}{Line}{Lines}
\theoremstyle{plain}

\theoremstyle{definition}

\theoremstyle{remark}

\definecolor{blue}{HTML}{49a0d4}
\definecolor{red}{HTML}{cc1100}
\definecolor{orange}{HTML}{cc7700}
\definecolor{gray}{HTML}{efefef}
\definecolor{darkgreen}{HTML}{228B22}
\definecolor{darkgray}{HTML}{757575}

\newcommand{\modelname}{\texttt{LapFlow}\xspace}
\newcommand{\lfm}{\texttt{LFM}\xspace}
\newcommand{\edifyimage}{\texttt{EdifyImage}\xspace}
\newcommand{\pyramidalflow}{\texttt{PyramidalFlow}\xspace}
\usepackage[textsize=tiny]{todonotes}

\newcommand{\mE}{{\mathbb E}}

\newcommand{\cN}{{\mathcal N}}

\newcommand{\bu}{{\mathbf u}}
\newcommand{\bv}{{\mathbf v}}
\newcommand{\bx}{{\mathbf x}}

\newcommand{\up}{{\rm Up}}
\newcommand{\down}{{\rm Down}}

\definecolor{grey}{rgb}{0.6,0.6,0.6}
\definecolor{lightgray}{rgb}{0.97,.99,0.99}
\title{Laplacian Multi-scale Flow Matching for Generative Modeling}

\author{Zelin Zhao \& Petr Molodyk \& Haotian Xue \& Yongxin Chen \thanks{The code is available at https://github.com/sjtuytc/gen.} \\
Georgia Institute of Technology\\
Atlanta, GA 30332, USA \\
\texttt{\{zelin,pmolodyk3,htxue.ai,yongchen\}@gatech.edu}
}

\iclrfinalcopy % Uncomment for camera-ready version, but NOT for submission.
\begin{document}
\maketitle

% Table tools
% https://tex.stackexchange.com/a/157400/72568
\newcolumntype{x}[1]{>{\centering\arraybackslash}p{#1}}
\newcolumntype{y}[1]{>{\raggedright\arraybackslash}p{#1}}
\newcolumntype{z}[1]{>{\raggedleft\arraybackslash}p{#1}}
\newcommand{\tablestyle}[2]{\setlength{\tabcolsep}{#1}\renewcommand{\arraystretch}{#2}\centering\footnotesize}

%Set up nice tables using the tabular package
%Make \toprule and \bottomrule thicker than \midrule
\setlength\heavyrulewidth{0.10em}
\setlength\lightrulewidth{0.05em}
\setlength\cmidrulewidth{0.03em}
\newcommand{\ra}[1]{\renewcommand{\arraystretch}{#1}}
\newcommand{\xue}[1]{\textcolor{blue}{[\textbf{xue}: #1]}}

\begin{abstract}
In this paper, we present Laplacian multiscale flow matching (LapFlow), a novel framework that enhances flow matching by leveraging multi-scale representations for image generative modeling. Our approach decomposes images into Laplacian pyramid residuals and processes different scales in parallel through a mixture-of-transformers (MoT) architecture with causal attention mechanisms. Unlike previous cascaded approaches that require explicit renoising between scales, our model generates multi-scale representations in parallel, eliminating the need for bridging processes. The proposed multi-scale architecture not only improves generation quality but also accelerates the sampling process and promotes scaling flow matching methods. Through extensive experimentation on CelebA-HQ and ImageNet, we demonstrate that our method achieves superior sample quality with fewer GFLOPs and faster inference compared to single-scale and multi-scale flow matching baselines. The proposed model scales effectively to high-resolution generation (up to 1024×1024) while maintaining lower computational overhead.
\end{abstract}

\section{Introduction}
\label{sec-introduction}
Generative modeling has seen remarkable progress in recent years, with diffusion models and flow matching approaches achieving state-of-the-art results in image synthesis~\citep{dhariwal2021diffusion,FlowMatchingGenerativeModeling,DiT}. These models have demonstrated impressive capabilities in generating realistic, high-quality, and diverse samples across various domains. However, as the demand for higher resolution and more complex content increases, scalability remains a significant practical challenge~\citep{VAR, PyramidFlow,EdifyImage}. As these methods typically generate the entire image at full resolution, they require substantial computational resources during training and inference phases. This motivates alternative approaches that maintain generation quality while potentially improving sampling efficiency and reducing inference overheads.

Multi-scale generation~\citep{LapGAN, PyramidFlow,EdifyImage,VAR,ho2022cascaded,NonUniformDiffusion} has emerged as a promising direction to address these scalability challenges. Existing multi-scale methods in both diffusion and flow matching paradigms, such as Cascaded Diffusion Models~\citep{ho2022cascaded}, EdifyImage~\citep{EdifyImage}, and Pyramidal Flow~\citep{PyramidFlow}, have demonstrated improved efficiency by generating images progressively from lower to higher resolutions. However, each of these approaches faces challenges that have limited their adoption compared to popular single-scale methods like DiT~\citep{DiT}. Cascaded Diffusion Models~\citep{ho2022cascaded} require training and maintaining separate networks for each resolution level, increasing implementation complexity. EdifyImage~\citep{EdifyImage}, while effective, operates directly in pixel space rather than latent space, resulting in significantly slower inference compared to latent-based methods. Pyramidal Flow~\citep{PyramidFlow} has demonstrated impressive results when fine-tuning DiTs for video generation, where they initialize their MM-DiT model from SD3 Medium~\citep{esser2024scaling} and fine-tune it afterwards, but its effectiveness when trained from scratch for image generation tasks remains less thoroughly explored in the literature.

\begin{figure}[!t]
  \centering
 \includegraphics[width=1.0\linewidth]{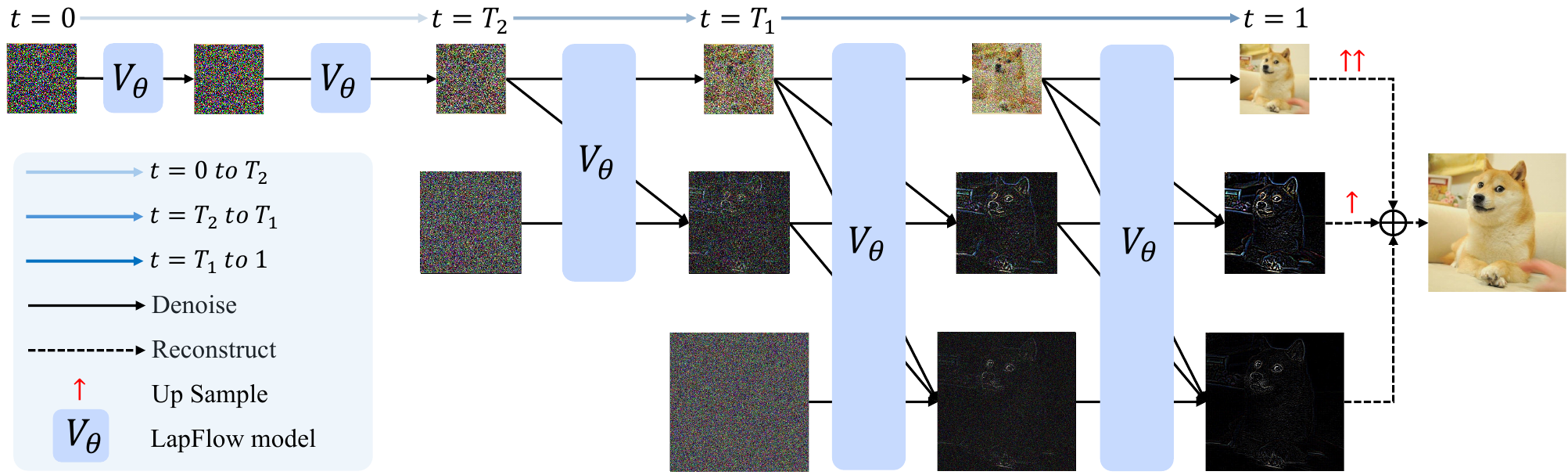}
 % \vspace{-5mm}
  \caption{\textbf{Multi-scale generation process of our model}. The proposed model follows a coarse-to-fine generation strategy across scales in a Laplacian pyramid. This figure demonstrates a three-level version of ours, where $T_2$, $T_1$ are two critical points defining three sampling segments for three scales. Starting from a random noise at $t=0$, our model first denoises the coarsest scale until $t=T_2$, then progressively conditions finer scales on completed coarser scales ($t=T_2$ to $T_1$ and $t=T_1$ to $1$). This causal structure ensures coherent image generation by maintaining hierarchical dependencies across scales, ultimately producing high-fidelity samples with both global consistency and fine details.}
  \label{fig-teaser}
% \vspace{-3mm}
\end{figure}

In this paper, we introduce \textit{Laplacian Multi-scale Flow Matching} (\modelname), enabling parallel modeling of multi-scale representations. As shown in~\Cref{fig-teaser}, the proposed model progressively generates Laplacian residuals through a unified model, and then reconstructs the full image through a hierarchical combination of these residuals following the Laplacian pyramid reconstruction process. Our approach enables parallel generation of multiple scales through a unified \textit{mixture-of-transformers} (MoT~\citep{MoT}) model with scale-specific modeling and shared-weight global attention. The model further employs causal attention mechanisms that enforce a natural information flow from lower to higher resolution scales, ensuring that finer details are coherently conditioned on broader structural elements while maintaining the hierarchical integrity of the image representation.

Through extensive experimentation on both CelebA-HQ~\citep{ProgressiveGan} and ImageNet~\citep{imageNet}, we demonstrate that our approach achieves superior sample quality compared to existing single-scale and multi-scale flow matching methods while requiring fewer compute (GFLOPs) during sampling. On CelebA-HQ, our method achieves an FID of $3.53$ at 256×256 resolution (compared to $5.26$ for LFM~\citep{LFM}), while our method keeps a strong performance scaling up to $1024\times 1024$. Comprehensive ablation studies on CelebA-HQ demonstrate the effectiveness of our key design choices, including the MoT architecture, causal masking, and noise scheduler, etc. For class-conditional generation on ImageNet, our approach outperforms both single-scale and multi-scale flow matching methods while maintaining lower computational requirements. 

% When trained for $7$M steps, our B/2 model achieves an FID of $4.12$ with CFG=$1.5$, outperforming the LFM baseline $4.46$ with fewer compute.

\textbf{Our contributions are summarized as follows:}
% \vspace{-2mm}
\begin{enumerate}
    \item We present a multi-scale flow matching framework that decomposes images into Laplacian pyramid representations, enabling joint modeling of different scale components.
    \item We introduce a specialized mixture-of-transformers (MoT) architecture with causal attention mechanisms that processes multiple scales simultaneously, significantly reducing inference compute while enforcing natural information flow between resolution levels. Through a time-weighted complexity analysis, we show that the effective attention cost of our progressive multi-scale design is theoretically lower than that of DiT~\citep{DiT}.
    \item We develop a progressive training strategy that optimizes different scales across distinct time ranges, allocating computational resources according to each scale's contribution.
\end{enumerate}

\section{Related Work}
\paragraph{Diffusion and flow matching.} Diffusion models~\citep{DiT,ddpm} have emerged as a dominant approach for generative modeling, achieving state-of-the-art results across various domains through iterative denoising processes. Despite their success, these models often require a large number of function evaluations during sampling, leading to computational inefficiency. Flow matching~\citep{FlowMatchingGenerativeModeling, albergo2022building, liu2022flow} is a simple yet effective framework for generative modeling, demonstrating strong performance across multiple domains, particularly in image generation~\citep{ma2024sit, esser2024scaling, liu2022flow, LFM} and video generation~\citep{PyramidFlow, deepmind2025veo2}. Instead of using noise or clean samples as learning objectives, it learns a velocity field that defines a deterministic transformation from the prior to the target distribution.

\paragraph{Multi-scale generation.}
Multi-scale approaches have a rich history in generative modeling, beginning with LapGAN~\citep{LapGAN}, which demonstrated the effectiveness of progressively generating high-resolution images from lower-resolution ones using GANs. This hierarchical concept was later adapted to diffusion models with cascaded architectures~\citep{ho2022cascaded, imagen, cascadedvideo}, which employ sequences of separate models to generate images of increasing resolution, each conditioned on lower-resolution outputs. Later on, Non-Uniform Diffusion Models~\citep{NonUniformDiffusion}, Relay Diffusion~\citep{relaydiffusion} and Pyramidal Flow~\citep{PyramidFlow} utilize renoising operations to bridge between resolution levels in a full generation cycle, while EdifyImage~\citep{EdifyImage} attenuates different frequency components at varying rates. However, these approaches typically require separate models or complex bridging mechanisms between scales, ignoring the use of causal relationships between scales~\citep{VAR}. In contrast, our \modelname framework employs a single mixture of transformers model that processes all scales simultaneously through causal attention mechanisms, eliminating the need for explicit bridging while allowing for more efficient generation.

\paragraph{Parameter-efficient architectures and auto-regressive generative models.} Parameter efficiency has become increasingly important in generative modeling as model sizes continue to grow. Mixture-of-Experts (MoE) approaches~\citep{shazeer2017outrageously, fedus2022switch} have emerged as a powerful paradigm for conditionally activating only relevant parameters, substantially improving computational efficiency. These concepts have been adapted to vision tasks through Mixture-of-Transformers (MoT)~\citep{MoT} architectures that enable specialized processing pathways. Concurrent developments in auto-regressive generative models~\citep{VAR,LlamaGen} have demonstrated the effectiveness of causal modeling. Our approach leverages flow matching with ODE-based parallel sampling, fundamentally differentiating it from auto-regressive methods that generate content sequentially and suffer from inherent parallelization constraints.
% \vspace{-3mm}
\section{Background}
\label{sec-flow-matching-backgrounds}
\paragraph{Flow matching formulations.}Flow matching~\citep{FlowMatchingGenerativeModeling} learns a time-dependent vector field $\bu_t$ that smoothly interpolates between a prior distribution $p_0$ and a data distribution $q$. We use $\bx_t$ to represent the evolving data sample over time. Flow matching directly optimizes a regression objective $\mathbb{E}_{t, q(\bx_1), p_t(\bx | \bx_1)} \left\| \bv_t(\bx) - \bu_t(\bx | \bx_1) \right\|^2$, minimizing the discrepancy between the learned vector field $\bv_t(\bx)$ and the conditional velocity of ground truth $\bu_t(\bx|\bx_1)$, called the Conditional Flow Matching (CFM) objective. Training a flow-matching model requires defining a valid interpolation path between $p_0$ and $q$.  Let $\bx_0 \sim p_0(\bx)$ be a noise variable, and $\bx_1 \sim q(\bx)$ be a data sample. We can derive the noisy data $\bx_t$ and the corresponding velocity field $\bu_t$ as $\bx_t = \alpha_t \bx_1 + \sigma_t \bx_0$ and $\bu_t(\bx_t | \bx_1) = \dot{\alpha}_t\bx_1 + \dot{\sigma}_t\bx_0$, where $\alpha_t$ and $\sigma_t$ are pre-defined coefficients for noise scheduling, while $\dot{\alpha}$ and $\dot{\sigma}$ are their temporal derivatives respectively. A simple yet effective choice~\citep{FlowMatchingGenerativeModeling, LFM} of $\alpha_t$ and $\sigma_t$ is linear interpolation: $\alpha_t = t, \sigma_t = 1-t$. Another popular probability path is the generalized variance preserving (GVP)~\citep{ma2024sit}:$\alpha_t = \sin{\left(\frac{1}{2}\pi t\right)}, \sigma_t = \cos{\left(\frac{1}{2}\pi t\right)}$. We choose to model flows in latent space~\citep{LFM}, aiming to reduce computational complexity and improve sample quality.
% \vspace{-3mm}
\paragraph{Latent flow matching.}While some flow matching methods attempt to directly model in pixel space~\citep{FlowMatchingGenerativeModeling}, LFM~\citep{LFM} and SiT~\citep{ma2024sit} propose to model in the latent space through a VAE, aiming to reduce computational complexity and improve sample quality.
% \vspace{-3mm}
\paragraph{Laplacian decomposition.}Denote the unknown data distribution (the image distribution) by $q(\bx_1)$. We define $\up$ as the nearest neighbor upsampling operation and $\down$ as the average pooling operation, both with a scaling factor of two for upsampling and downsampling, respectively. Given a data $\bx_1\sim q$, we decompose it into three residuals via the Laplacian decomposition (we formulate three scales here for simplicity, while it can easily generalize to other numbers of scales): 
\begin{equation}
\bx_1^{(2)} = \down(\down(\bx_1)),\quad
\bx_1^{(1)} = \down(\bx_1) - \up(\bx_1^{(2)}),\quad
\bx_1^{(0)} = \bx_1 - \up(\down(\bx_1)).
\end{equation}
In this work, we use the superscript $2$ to denote the smallest scale and use $0$ to denote the largest scale. Given the three scales' representation, the full data $\bx_1$ can be reconstructed from three residuals as
\begin{equation}
    \bx_1 = \bx_1^{(0)} + \up(\bx_1^{(1)}) + \up(\up(\bx_1^{(2)})).
\label{eq-laplacian-decomposition-combine}
\end{equation}
% Similarly, we also build a Laplacian pyramid for the pure noise $\bx_0$, and we denote Laplacian noises to be $\{\bx_0^{(0)}, \bx_0^{(1)}, \bx_0^{(2)}\}$.

\section{Methodology}
\vspace{-5mm}
% Our method is based on the flow matching framework~\citep{FlowMatchingGenerativeModeling}, where we will review the basics in~\Cref{sec-flow-matching-backgrounds}. After this, we introduce our \modelname model, which builds a flow-matching model based on scalable representations.

\begin{algorithm}[tb]
   \caption{\modelname Training}
    \begin{algorithmic}[1]
    \STATE \textbf{Input:} image dataset $D$, number of epochs $M$, critical time points $0\triangleq T_3 < T_2 < T_1 < 1$
    % \STATE \textbf{Output:} Trained multiscale flow model $\bv_t^{(0)}, \bv_t^{(1)}, \bv_t^{(2)}$
    \STATE Initialize weights $\theta$ of the~\modelname model $V_\theta$
    \FOR{epoch = 1 to $M$}
        \FOR{each image $\bx_1$ in $D$}
            \STATE Sample a random noise image $\bx_0$ and generate Laplacian pyramid $\{\bx_0^{(0)}, \bx_0^{(1)}, \bx_0^{(2)}\}$
            \STATE Sample stage $s \sim \mathcal{U}\{0, 1, 2\}$ \label{line-stage-sample}, and sample time $t \sim \mathcal{U}[T_{s+1}, 1]$ \label{line-time-sample}
            \STATE Compute noisy image through~\Cref{eq-general-bxtk} for $k \geq s$ and $k\in\{0,1,2\}$
            \STATE Compute velocity target through~\Cref{eq-gt-velocity} for $k\geq s$ and $k\in\{0,1,2\}$
            \STATE Forward the model to get the predictions through~\Cref{eq-model-prediction} \label{line-model-forward}
            \STATE Calculate the multiscale velocity loss $\mathcal{L}_{mv}$ as~\Cref{eq-multiscale-velocity-loss} \label{line-ms-loss}
            \STATE Run back-propagation and update network parameters
        \ENDFOR
    \ENDFOR
    \STATE \textbf{Return:} trained \modelname model $V_\theta$
    \end{algorithmic}
\label{alg-ours-train}
\end{algorithm}
\makeatletter
\setcounter{ALC@unique}{0}
\makeatother
\begin{algorithm}[tb]
\caption{\modelname Sampling}
\label{our-sample}
\begin{algorithmic}[1]
\STATE \textbf{Input:} trained flow model $V_\theta$, a largest-scale random noise $\bx_0$, critical time points $T_1$ and $T_2$
% \STATE \textbf{Output:} Largest resolution sample
\STATE Get Laplacian pyramid $\{\bx_0^{(0)}, \bx_0^{(1)}, \bx_0^{(2)}\}$ from $\bx_0$ \label{line-laplacian-gaussian}
\STATE $\{\hat{\bx}_{T_2}^{(2)}\} = \operatorname{ODEINT}(V_\theta, \{0, T_2\}, \{\bx_0^{(2)}\})$ \label{line-infer-step-1}
\STATE $\{\hat{\bx}_{T_1}^{(1)}, \hat{\bx}_{T_1}^{(2)}\} = \operatorname{ODEINT}(V_\theta, \{T_2, T_1\}, \{\sigma_{T_2}^{(1)}\bx_0^{(1)}, \hat{\bx}_{T_2}^{(2)}\})$ \label{line-infer-step-2}
\STATE $\{\hat{\bx}_{1}^{(0)}, \hat{\bx}_{1}^{(1)}, \hat{\bx}_1^{(2)}\} = \operatorname{ODEINT}(V_\theta, \{T_1, 1\}, \{\sigma_{T_1}^{(0)}\bx_0^{(0)}, \hat{\bx}_{T_1}^{(1)}, \hat{\bx}_{T_1}^{(2)}\})$ \label{line-infer-step-3}
\STATE $\hat{\bx}_{1} = \hat{\bx}_{1}^{(0)} + \up(\hat{\bx}_{1}^{(1)}) + \up(\up(\hat{\bx}_{1}^{(2)}))$
\STATE \textbf{Return:} the largest-scale sample $\hat{\bx}_{1}$ 
\end{algorithmic}
\label{alg-ours-sample}
\end{algorithm}

\subsection{Multi-scale Noising Process}

We propose learning the components of three scales at different speeds. We identify two critical time points $T_{1}$ and $T_{2}$ as two hyperparameters, where $0\triangleq T_3 < T_2 < T_1 < 1$. In addition, we denote $T_0=1$ and $T_3=0$ for convenience of notation. For the $k$-th scale, we train the model from $t=T_{k+1}$ to $t=1$, which means that the model of larger scales is trained for a shorter period of time. In a time step $t$ (note that $T_{k+1}\leq t \leq 1$ ensured by our progressive multi-stage training strategy, which will be presented in~\Cref{sec-progressive-multi-stage-training}), we construct the noisy data of scale $k$ via a weighted sum of $\bx_0^{(k)}$ and $\bx_1^{(k)}$:
\begin{equation}
\bx_t^{(k)} = \alpha_t^{(k)} \bx_1^{(k)} + \sigma_t^{(k)} \bx_0^{(k)},
\label{eq-general-bxtk}
\end{equation}
where $\{\alpha_t^{(k)}, \sigma_t^{(k)}\}$ are pre-defined per-scale coefficients for data and noise. The interpolation can be conducted through a linear path or a GVP path by modifying the coefficients:
\begin{subequations}
\begin{equation}
\text{Linear:} \quad \alpha_t^{(k)} = \frac{t - T_{k+1}}{1 - T_{k+1}}, \quad \sigma_t^{(k)} = 1 - t,
\end{equation}
\begin{equation}
\text{GVP:} \quad \alpha_t^{(k)} = \sin\left(\frac{\pi (t - T_{k+1})}{2(1 - T_{k+1})}\right), \quad \sigma_t^{(k)} = \cos\left(\frac{\pi}{2}t\right).
\end{equation}
\label{eq-scheduler-ours}
\end{subequations}
This multi-scale flow satisfies two key properties: (1) At the starting time $t=T_{k+1}$ for scale $k$, the noisy data contains only the scaled noise component, as $\alpha_{T_{k+1}}^{(k)}=0$ results in $\bx_{T_{k+1}}^{(k)}=\sigma_{T_{k+1}}^{(k)}\bx_0^{(k)}$; (2) At the ending time $t=1$, the representation $\bx_1^{(k)}$ converges to the clean Laplacian residual of scale $k$, completing the generation process. The proposed flow provides the following multi-scale velocity:
\begin{equation}
\bu_t^{(k)}(\bx_t^{(k)} | \bx_1^{(k)})= \dot{\alpha}_{t}^{(k)}\bx_1^{(k)} + \dot{\sigma}_{t}^{(k)} \bx_0^{(k)},
\label{eq-gt-velocity}
\end{equation}
where $\dot{\alpha}_{t}^{(k)}$ and $\dot{\sigma}_{t}^{(k)}$ are temporal derivatives of $\alpha_{t}^{(k)}$ and $\sigma_{t}^{(k)}$ respectively.

\subsection{Progressive Multi-stage Training and Training Targets}
\label{sec-progressive-multi-stage-training}
The training algorithm is presented in~\Cref{alg-ours-train}. We adopt a progressive training scheme~\citep{ProgressiveGan}, which trains different-scale models for different time ranges. We introduce the concept of \textit{stage} (denoted by $s\sim U([0, 1, 2])$, where in different stages, we train different scales of models. In training stage $s\in \{0, 1, 2\}$, we train all scales $k$ such that $k \geq s$ (noting that smaller $k$ values represent higher resolutions, with $k=0$ being the highest resolution and $k=2$ being the lowest). After sampling $s$, the timestep $t$ is then sampled from the range $[T_{s+1}, 1]$ (\Cref{line-time-sample}). As a result, the smallest scale ($k=2$) is trained throughout the time range $[0,1]$, the mid-resolution scale ($k=1$) is trained across $[T_2, 1]$, and the highest resolution scale ($k=0$) is trained only during $[T_1, 1]$.

Our model takes noisy multi-scale data $\{\bx_t^{(k)}\}_{k=2}^s$ as input, and outputs corresponding multi-scale velocities $\{\bv_t^{(k)}\}_{k=2}^s$. Both input and output consist of scales from the smallest scale ($k=2$) to the current scale ($k=s$), while they are conditioned on the same time $t$. Let $V_\theta$ be the~\modelname model (see~\Cref{sec-multiscale-transformer-arch} for a detailed architecture), the model prediction (\Cref{line-model-forward}) is formulated as:
\begin{equation}
\{\bv_t^{(k)}\}_{k=2}^s = V_\theta\left(\{\bx_t^{(k)}\}_{k=2}^s \right).
\label{eq-model-prediction}
\end{equation}
% \begin{equation}
% \{\bv_t^{(2)}(\bx_t^{(2)}), \bv_t^{(1)}(\bx_t^{(2)},\bx_t^{(1)}), \bv_t^{(0)}(\bx_t^{(2)},\bx_t^{(1)},\bx_t^{(0)})\} = \texttt{\modelname}\left(\{\bx_t^{(2)},\bx_t^{(1)},\bx_t^{(0)}\} \right).
% \label{eq-model-prediction}
% \end{equation}
We apply a multiscale conditioning flow matching loss (\Cref{line-ms-loss}) as:
\begin{equation}
\mathcal{L}_{mv} = \sum_{k=2}^{s}w_{k}\mathbb{E}_{t, q(\bx_1^{(k)}), p_t(\bx_t^{(k)} | \bx_1^{(k)})}\|\bv_t^{(k)} - \bu_t^{(k)}(\bx_t^{(k)}|\bx_1^{(k)}) \|^2,
\label{eq-multiscale-velocity-loss}
\end{equation}
% \begin{equation}
% \mathcal{L}_{mv} = \sum_{k=2}^{s}w_{k}\mathbb{E}_{t, q(\bx_1^{(k)}), p_t(\bx | \bx_1^{(k)})}\|\bv_t^{(k)}(\bx) - \bu_t^{(k)}(\bx|\bx_1^{(k)}) \|^2,
% \label{eq-multiscale-velocity-loss}
% \end{equation}
where $w_k$ is a balancing factor for scale $k$'s loss, where in practice we set $w_k=1$ for simplicity.

\subsection{Multi-scale Sampling Process}
\label{sec-multi-scale-sampling-process}
We present our sample algorithm in~\Cref{alg-ours-sample}, which is used to generate images with the highest resolution. The sampling algorithm starts with a Laplacian of noise (\Cref{line-laplacian-gaussian}). We then leverage the ODE solving function $\operatorname{ODEINT}$, which takes three arguments (a model $V_\theta$, initial and end timesteps $\{t_i, t_e\}$, and an initial condition $\bx_{t_i}$) as input and outputs the variable $\bx_{t_e}$ corresponding to the end timestep $t_e$. A popular implementation of $\operatorname{ODEINT}$ that we use is the $\operatorname{torchdiffeq}$ library~\citep{chen2018neuralode}. This general process can be formally expressed as $\bx_{t_e}=\operatorname{ODEINT}(V_\theta, \{t_i, t_e\}, \bx_{t_i})$.

We first denoise the smallest scale from $t=0$ to $T_2$ to get the partially denoised state $\bx_{T_2}^{(2)}$ (\Cref{line-infer-step-1}). Then we continue the denoising procedure during $T_2\le t < T_1$, where we denoise the mid-scale and smallest scale in parallel (\Cref{line-infer-step-2}). In this step, we leverage the fact that when $t=T_2$, the mid-resolution noisy state becomes $\sigma_{T_2}^{(1)}\bx_0^{(1)}$, while the smallest noisy state becomes $\hat{\bx}_{T_2}^{(2)}$ according to~\Cref{eq-general-bxtk}. Therefore, this step would take an initial condition $\{\sigma_{T_2}^{(1)}\bx_0^{(1)}, \hat{\bx}_{T_2}^{(2)}\}$. Similarly, the last step would take $\{\sigma_{T_1}^{(1)}\bx_0^{(0)}, \hat{\bx}_{T_1}^{(1)}, \hat{\bx}_{T_1}^{(2)}\}$ as the initial condition. We finally evaluate the model from $T_1$ to $1$ (\Cref{line-infer-step-3}), and these denoised residuals are then combined into full-scale data via~\Cref{eq-laplacian-decomposition-combine}.

\begin{figure}[t]
  \centering
 \includegraphics[width=1.0\linewidth]{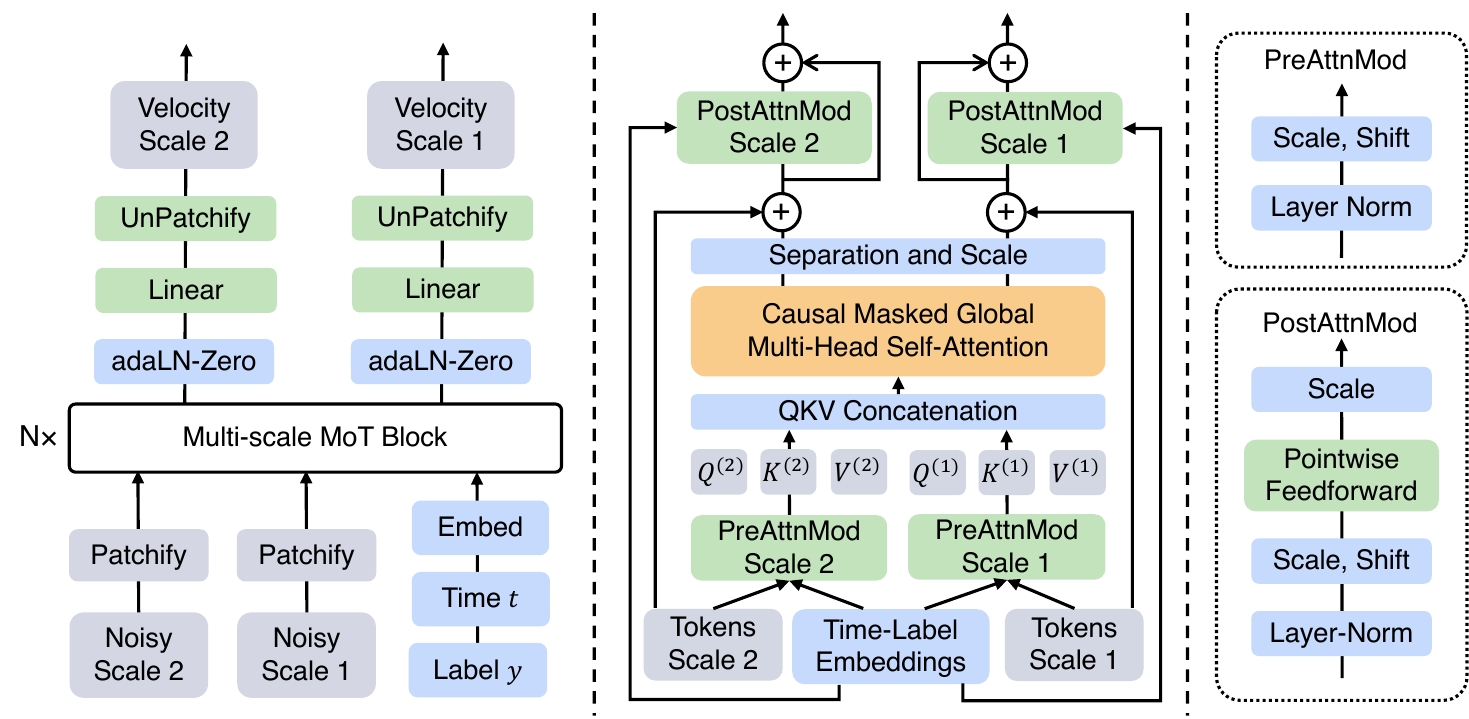}
  \caption{(\textbf{Left:}) Schematic of the~\modelname model $V_\theta$. The multi-scale transformer takes multi-scale noisy states as input, conditioned on time and label, and predicts velocities for each input scale. While the model can take an arbitrary number of scales as input, we show two here for simplicity. (\textbf{Middle:}) Details of one multi-scale MoT block. We use separate QKVs for different scales, while the attention is computed globally. Furthermore, we adopt a mask to enforce causal relationships across scales. (\textbf{Right:}) Details of scale-specific \texttt{PreAttnMod} and \texttt{PostAttnMod} modules~\citep{DiT}, where each \texttt{PostAttnMod} module includes a feedforward network (FFN).}
  \label{fig-model}
% \vspace{-4mm}
\end{figure}

\subsection{Multi-scale DiT with Mixture-of-Transformers}
\label{sec-multiscale-transformer-arch}
We present the~\modelname model, which is a multi-scale diffusion transformer (DiT) with Mixture-of-Transformers (MoT~\citep{MoT}). The complete model architecture of~\modelname is shown in~\Cref{fig-model}. As formulated in~\Cref{eq-model-prediction}, our \modelname architecture is designed with flexible input-output capabilities: it can process noisy states of any subset of scales (from one to three scales) as input, and will correspondingly output velocity predictions for exactly the same scales that were provided as input. This flexibility enables the model to handle different stages of the generation process where varying numbers of scales are active. We inherit best practices from DiT~\citep{DiT}, including patchify, feature modulations, and in-context conditioning. In addition, we utilize MoT~\citep{MoT} to enable scale-specific processing (e.g., pre-attention and post-attention modulations) with global multi-head self-attention over all scales. In the global attention mechanism, \modelname implements a causal masking strategy that enforces a unidirectional information flow from smaller scales (lower resolutions) to larger scales (higher resolutions). 
\vspace{-1mm}
\paragraph{Model schematic (left of~\Cref{fig-model}).}
The input noisy states at each scale are first converted into sequences of tokens using the $\texttt{Patchify}$ operation~\citep{DiT}. For each scale, sine-cosine positional embeddings~\citep{ViT} are added to its corresponding tokens. Conditions such as time $t$ and label $y$ are added as additional tokens to the input sequence (in-context conditioning~\citep{DiT}, enabling flexible conditioning without architectural modifications. The input tokens are then processed by $N$ multi-scale MoT blocks, followed by an adaptive Layer Norm layer with zero initialization (\texttt{adaLN-Zero}~\citep{MoT}). Finally, a linear decoder and an $\texttt{UnPatchify}$ (rearrange) process are applied to get velocity predictions. 

\paragraph{Multi-scale MoT block (middle and right of~\Cref{fig-model}).} For any scale $k\in \{s, ..., 2\}$ (where $s$ is the current training stage), we use $\alpha_1^{(k)}, \alpha_2^{(k)}, \beta_1^{(k)}, \beta_2^{(k)}, \gamma_1^{(k)}$ and $ \gamma_2^{(k)}$ to denote parameters for scales and shifts regressed from conditional embeddings, following DiT convention~\citep{DiT}. Firstly, we apply a pre-attention modulation process (\texttt{PreAttnMod}) that includes both scale-and-shift operations through factors $\gamma_1^{(k)}$ and $\beta_1^{(k)}$, as well as scale-specific projections to obtain queries, keys, and values (QKVs). Let $z^{(k)}$ be the tokens after the scale-and-shift, and the scale-specific QKVs are:
\begin{equation}
Q^{(k)} = z^{(k)} W_Q^{(k)}, K^{(k)} = z^{(k)} W_K^{(k)}, V^{(k)} = z^{(k)} W_V^{(k)},
\label{eq-qkvs-each-scale}
\end{equation}
where $W_Q^{(k)}, W_K^{(k)}$ and $W_V^{(k)}$ are scale-specific weight matrices. Then, a masked global multi-head self-attention is applied to concatenations of QKVs from all scales. The single-head formulation is:
\begin{equation}
\texttt{MaskedGlobalAttn}(Q, K, V) = \texttt{Softmax}\left( \frac{QK^\top}{\sqrt{d}} + M_c \right) V,
\label{eq-global-attention}
\end{equation}
where \( Q = [Q^{(s)}; \ldots; Q^{(2)}] \), \( K = [K^{(s)}; \ldots; K^{(2)}] \), \( V = [V^{(s)}; \ldots; V^{(2)}] \) are concatenations of scale-specific QKVs, and $d$ is the feature dimension. We use a block causal mask~\citep{VAR} $M_c$ to ensure that each scale $k$ can only attend to scales smaller than or equal to $k$ ($k' \geq k$). The output of the global attention, a unified sequence of token representations, is subsequently partitioned according to scale, yielding separate feature sequences for each scale. Each scale's feature is then scaled by $\alpha_1^{(k)}$. Finally, we apply the $\texttt{PostAttnMod}$ module, which consists of layer normalization, a scale-and-shift operation (using $\gamma_2^{(k)}$ and $\beta_2^{(k)}$), a point-wise feed-forward network (FFN), and a final scaling by $\alpha_2^{(k)}$. Residual connections are applied to connect scale-wise representations.

% \vspace{-10mm}
% \paragraph{Theoretical complexity analysis.}
% Beyond the empirical GFLOPs reported in our tables, we provide a formal time-weighted complexity analysis of the multi-scale MoT attention mechanism (Appendix~\ref{app:mot-complexity}). Following the Laplacian decomposition, the three spatial scales contain $\{N,\, N/4,\, N/16\}$ latent tokens, and these scales are active over progressively shorter segments of the ODE time domain. Under the default temporal partition $(T_2, T_1) = (0.33, 0.67)$, the token counts in the three corresponding segments are
% $N_2 = \tfrac{N}{16}$,
% $N_1 = \tfrac{5N}{16}$,
% and $N_0 = \tfrac{21N}{16}$.
% Averaging the quadratic attention cost over the three equally sized segments yields the per-layer time-weighted complexity
% $
% \text{Cost}_{\text{MoT}} \approx\;
% 0.61\,N^2 d,
% $
% which corresponds to a theoretical attention reduction of roughly $1.6\times$ compared to the $\mathcal{O}(N^2 d)$ cost of a single-scale DiT~\citep{DiT}.
% \vspace{-4mm}
\section{Experiments}
\label{sec-experiments}
% \vspace{-3mm}
\subsection{Experimental Setup}
\paragraph{Datasets.} We conduct experiments on two datasets: CelebA-HQ~\citep{ProgressiveGan} and ImageNet~\citep{imageNet}. We use the term ``resolution'' to indicate the largest-scale resolution we are trying to generate (which is the resolution of the output to our sampling algorithm~\Cref{alg-ours-sample}). For CelebA-HQ, we experiment with resolutions $256\times256$, $512\times512$, and $1024\times1024$. For ImageNet, the resolution is set at $256\times 256$ due to limited resources. CelebA-HQ is unconditional, while ImageNet is conditioned on class labels. The VAE down-sample factor is eight, so the largest latent is $32\times 32$. We experiment with varied classifier-free guidance (CFG)~\citep{ho2022classifier} on ImageNet, while CFG is not enabled by default.
% \vspace{-3mm}
\paragraph{Baselines.}We benchmark and compare against several flow matching baselines on CelebA-HQ, including LFM~\citep{LFM}, which is a single-scale baseline. We also compare with a recent multi-scale method called~\pyramidalflow~\citep{PyramidFlow} in the task of image generation. Besides, we compare with~\edifyimage, which is an image-space method using multi-scale generation. We re-implement~\edifyimage and~\pyramidalflow under the flow matching framework for comparison. On ImageNet, we further evaluate our model against a broad range of recent works in both diffusion and flow matching paradigms. All models use DiT-L/2 for CelebA-HQ, and we experiment with DiT-B/2 and DiT-XL/2 for ImageNet. We utilize the Dormand-Prince method (dopri5) implemented in the torchdiffeq library~\citep{chen2018neuralode}. We adopt the Fréchet inception distance~\citep{heusel2017gansFID} with $50K$ images as the main metric.

\begin{table}[t]
\centering
\caption{\textbf{Evaluations on CelebA-HQ.} For all multi-scale methods (EdifyImage~\citep{EdifyImage}, Pyramidal Flow~\citep{PyramidFlow}, Relay Diffusion~\citep{relaydiffusion}, and ours), we report averaged GFLOPs and time during inference. }
\resizebox{\textwidth}{!}{%
\begin{tabular}{@{}l|c|c|c|c|c|c|c|c|c@{}}
\toprule
Methods & Backbone & Resolution & Space & FID$\downarrow$ & Recall$\uparrow$ & NFE & Time (s) & Params & GFLOPs \\ \midrule
\multicolumn{10}{c}{\textit{\textbf{Standard Resolution (256)}}} \\ \midrule
FM~\citep{FlowMatchingGenerativeModeling} & U-Net & $256\times256$ & Image & 7.34 & - & 128 & - & - & - \\
LDM~\citep{LDM} & U-Net & $256\times256$ & Latent & 5.11 & 0.49 & 50 & 2.90 & 150M & 10.2 \\
LFM~\citep{LFM} & DiT-L/2 & $256\times256$ & Latent & 5.26 & 0.46 & 89 & 1.70 & 541M & 22.1 \\
Pyramidal Flow~\citep{PyramidFlow} & DiT-L/2 & $256\times256$ & Latent & 11.20 & 0.48 & 90 & 1.85 & 544M & 14.2 \\
EdifyImage~\citep{EdifyImage} & DiT-L/2 & $256\times256$ & Image & 7.62 & 0.47 & 95 & 2.10 & 602M & 28.9 \\
\rowcolor{gray}
Ours & DiT-L/2 & $256\times256$ & Latent & 3.53 & 0.53 & 80 & 1.51 & 612M & 16.5 \\ \midrule
\multicolumn{10}{c}{\textit{\textbf{High Resolution (512 and 1024)}}} \\ \midrule
LFM~\citep{LFM} & DiT-L/2 & $512\times512$ & Latent & 6.35 & 0.41 & 93 & 2.90 & 544M & 43.5 \\
\rowcolor{gray}
Ours & DiT-L/2 & $512\times512$ & Latent & 4.04 & 0.47 & 85 & 2.60 & 680M & 41.7 \\
LFM~\citep{LFM} & DiT-L/2 & $1024\times1024$ & Latent & 8.12 & 0.34 & 100 & 4.20 & 544M & 154.8 \\
\rowcolor{gray}
Ours & DiT-L/2 & $1024\times1024$ & Latent & 5.51 & 0.39 & 94 & 3.30 & 680M & 148.2 \\ \bottomrule
\end{tabular}%
}
\label{tab-celebahq-results}
\end{table}
% \vspace{-4mm}
\begin{figure}[t]
  \centering
 \includegraphics[width=1.0\linewidth]{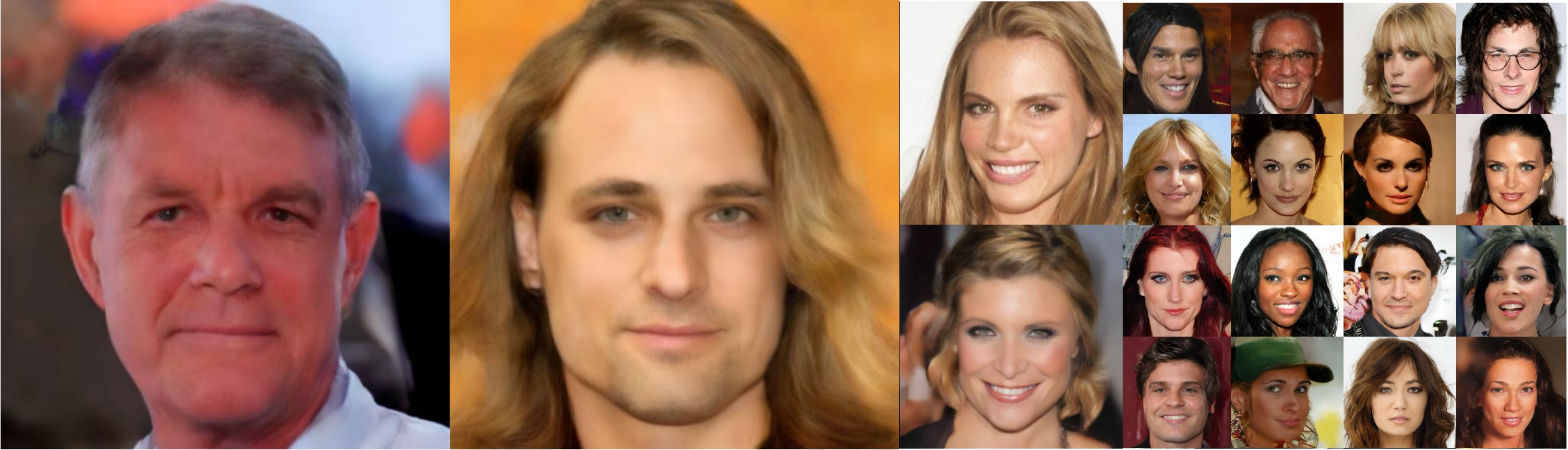}
 % \vspace{-4mm}
  \caption{Qualitative results on CelebA-HQ 1024 (left two), 512 (middle two), and 256 (right).}
  \label{fig-celeba-res}
% \vspace{-5mm}
\end{figure}
\begin{table}[t!]
    \caption{\textbf{Ablation studies on CelebA-HQ.} We analyze various design choices, including VAE architecture, mixture-of-Transformer (MoT) design, noise schedule, and other modeling configurations. We report FID-50K scores on $256 \times 256$ resolution images. Our default settings are marked in \colorbox{gray}{gray}.}
    \label{tab:ablation}
    \vspace{-0mm}
    \begin{minipage}{0.48\textwidth}
        \centering
        \subcaption{\textbf{VAE choices.} EQVAE~\citep{EQVAE} benefits our method over SDVAE~\citep{LDM} but it cannot improve LFM~\citep{LFM}.}\label{subtab-vae}
        \vspace{-2mm}
        \tablestyle{0pt}{1.08}
        \begin{tabular}{x{30pt}| x{38pt} x{38pt} x{38pt} x{38pt}}\toprule
        Method &LFM (SDVAE) &LFM (EQVAE) & Ours (SDVAE) & \cellcolor[HTML]{efefef} Ours (EQVAE)\\\midrule
        FID &5.26 &7.77 &4.37 &\cellcolor[HTML]{efefef} \textbf{3.53} \\
        \bottomrule
        \end{tabular}
    \end{minipage}
    \hspace{1.5mm}
    \centering
    \begin{minipage}{0.48\textwidth}
    \subcaption{\textbf{MoT design.} Comparison between separate weights models and our proposed MoT model, the MoT can reduce the GFLOPs by selecting experts.}\label{subtab-mot}
    \vspace{-2mm}
    \centering
    \tablestyle{0pt}{1.08}
    \begin{tabular}{x{40pt}| x{68pt} x{58pt}}\toprule
    Metric & Separate Model &\cellcolor[HTML]{efefef} MoT \\\midrule
    GFLOPs & 38.9 &\cellcolor[HTML]{efefef} \textbf{16.5} \\
    FID &3.60 &\cellcolor[HTML]{efefef}\textbf{3.53} \\ 
    \bottomrule
    \end{tabular}
    \end{minipage}\\
    \vspace{0mm}
    \begin{minipage}{0.48\textwidth}
        \subcaption{\textbf{Causal masking.} ``None'' means no mask is used, ``Self'' means each scale is only attended on itself.}\label{subtab-causal}
        \vspace{-2mm}
        \centering
        \tablestyle{0pt}{1.08}
        \begin{tabular}{x{45pt}| x{48pt} x{48pt} x{48pt}}\toprule
        Mask &None &Self &\cellcolor[HTML]{efefef}Causal \\\midrule
        FID &3.91 &5.19 &\cellcolor[HTML]{efefef}\textbf{3.53} \\
        \bottomrule
        \end{tabular}
    \end{minipage} 
    \hspace{1.5mm}
    \begin{minipage}{0.48\textwidth}
        \centering
        \subcaption{\textbf{Critical time point $T$.} Compare $T$ values, which determines the temporal segments of two scales.}\label{subtab:coupling}
        \vspace{-2mm}
        \tablestyle{0pt}{1.08}
       \begin{tabular}{x{34pt}| x{38pt} x{38pt} x{38pt} x{38pt}}\toprule
        $T$ &0.1 &0.2 &\cellcolor[HTML]{efefef}0.5 &0.9 \\\midrule
        FID &5.12 &4.37 &\cellcolor[HTML]{efefef}\textbf{3.53} &4.92 \\
        \bottomrule
        \end{tabular}
    \end{minipage} \\
    \vspace{0mm}
    \begin{minipage}{0.48\textwidth}
        \subcaption{\textbf{Learning rate decay final LR.} The final learning rate of $1\times 10^{-6}$ achieves optimal performance, with both higher and lower values degrading results.}\label{subtab:optimization}
        \vspace{-2mm}
        \centering
        \tablestyle{0pt}{1.08}
        \begin{tabular}{x{34pt}| x{38pt} x{38pt} x{38pt} x{38pt}}\toprule
        Final LR & $2\times 10^{-4}$ & $1\times 10^{-5}$ &\cellcolor[HTML]{efefef}$1\times 10^{-6}$ &$1\times 10^{-7}$ \\\midrule
        FID &4.12 &3.81 &\cellcolor[HTML]{efefef} \textbf{3.53} & 4.82 \\
        \bottomrule
        \end{tabular}
    \end{minipage}
    \hspace{1.5mm}
      \begin{minipage}{0.48\textwidth}
        \centering
        \subcaption{\textbf{Noise schedule.} The simple linear schedule outperforms the GVP in our model, refer to \Cref{eq-scheduler-ours}.}\label{subtab-noise}
        \vspace{-2mm}
        \tablestyle{0pt}{1.08}
        \begin{tabular}{x{36pt}| x{38pt} x{38pt} x{38pt} x{38pt}}\toprule
        Noise Schedule &LFM (Linear) & LFM (GVP) & \cellcolor[HTML]{efefef} Ours (Linear) &Ours (GVP) \\\midrule
        FID & 5.26 & 4.39 & \cellcolor[HTML]{efefef} \textbf{3.53} & 4.10 \\
        \bottomrule
        \end{tabular}
    \end{minipage}\\
    % \vspace{0mm}
    \begin{minipage}{0.48\textwidth}
        \centering
        \subcaption{\textbf{Number of scales.} Two-scale representation achieves optimal performance in $256\times 256$, with more scales increasing complexity without further gains.}\label{subtab-num_scales}
        \vspace{-2mm}
        \tablestyle{0pt}{1.08}
        \begin{tabular}{x{34pt}| x{38pt} x{38pt} x{38pt} x{38pt}}\toprule
        \# of Scales & 1 (LFM) & \cellcolor[HTML]{efefef} 2 (Ours) & 3 (Ours)& 4 (Ours)\\ \midrule
        FID & 5.26 & \cellcolor[HTML]{efefef} \textbf{3.53} &3.59 & 5.12 \\
        \bottomrule
        \end{tabular}
    \end{minipage}
    \hspace{1.5mm}
\begin{minipage}{0.48\textwidth}
    \subcaption{\textbf{Modeling Space.} Latent space modeling significantly outperforms image space for both approaches, with our method showing substantial gains in both.}
    \label{subtab:nfe}
    \vspace{-2mm}
    \centering
    \tablestyle{0pt}{1.08}
    \begin{tabular}{x{34pt}| x{38pt} x{38pt} x{38pt} x{38pt}}\toprule
    Method & LFM (Image) & LFM (Latent) & Ours (Image) & \cellcolor[HTML]{efefef}Ours (Latent) \\\midrule
    FID & 11.58 & 5.26 & 8.63 & \cellcolor[HTML]{efefef}\textbf{3.53} \\
    \bottomrule
    \end{tabular}
\end{minipage}
    \vspace{-4mm}
\end{table}
\begin{table}[t]
\centering
       \caption{\textbf{Results of class-conditional generation on ImageNet $256\times256$.}}
       \vspace{1mm}
       \resizebox{\textwidth}{!}{%
     \begin{tabular}{@{}l|c|c|c|c|c|c|c@{}}
       \toprule
       Methods & Backbone & Training Steps & CFG & FID & NFE & Time (s) & GFLOPs \\ \midrule
       DiT~\citep{DiT} & DiT-B/2 & $600$K & $\times$ & 41.27 & 250 & 1.87 & 5.6 \\
       LFM~\citep{LFM} & DiT-B/2 & $600$K & $\times$ & 48.29 & 89 & 1.70 & 5.6 \\
       Pyramidal Flow~\citep{PyramidFlow} & DiT-B/2 & $600$K & $\times$ & 39.40 & 90 & 1.74 & 5.7 \\
       \rowcolor{gray}
       Ours & DiT-B/2 & $600$K & $\times$ & \textbf{36.50} & 80 & 1.25 & 4.9 \\
       \midrule
       DiT~\citep{DiT} & DiT-XL/2 & $600$K & $\times$ & 19.50 & 250 & 3.50 & 29.1 \\
      LFM~\citep{LFM} & DiT-XL/2 & $600$K & $\times$ & 28.37 & 89 & 3.32 & 29.1 \\
    Pyramidal Flow~\citep{PyramidFlow} & DiT-XL/2 & $600$K & $\times$ & 17.10 & 90 & 4.02 & 29.9 \\
       \rowcolor{gray}
       Ours & DiT-XL/2 & $600$K & $\times$ & \textbf{14.38} & 80 & 2.85 & 20.5 \\ \midrule
      LFM~\citep{LFM} & DiT-B/2 & $7$M & $\times$ & 20.38 & 89 & 1.70 & 5.6 \\
      LFM~\citep{LFM} & DiT-B/2 & $7$M & 1.5 & 4.46 & 89 & 1.70 & 5.6 \\
     \rowcolor{gray} Ours & DiT-B/2 & $7$M & $\times$ & 18.55 & 80 & 1.25 & 4.9 \\
    \rowcolor{gray} Ours & DiT-B/2 & $7$M & 1.5 & \textbf{4.12} & 80 & 1.25 & 4.9 \\
       \bottomrule
       \end{tabular}
       }
\label{tab-imagenet-results}
\end{table}
\begin{figure}[t]
  \centering
 \includegraphics[width=1.0\linewidth]{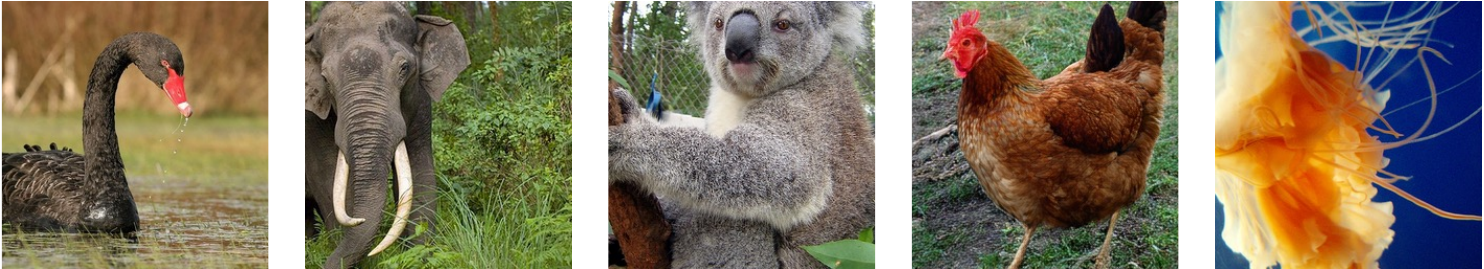}
 \vspace{-4mm}
  \caption{Qualitative results on ImageNet $256\times256$ using our trained B/2 model with CFG=$1.5$.}
  \label{fig-imagenet-sampling-res}
\vspace{-2mm}
\end{figure}
\paragraph{VAE.} While a popular choice of VAE is the SDVAE~\citep{LDM}, a recent work called EQVAE~\citep{EQVAE} introduces an equivariant regularization technique that enhances latent space structure by preserving scaling transformations. In practice, we find that EQVAE greatly benefits our multi-scale approach because it offers equivalent representations across scales. Since EQVAE is only trained on $256\times 256$, we use SDVAE instead for higher resolutions.

% \vspace{-3mm}
\paragraph{Hyperparameters and training settings.} For experiments that generate resolutions $256\times 256$, we use \textbf{two levels} in all multi-scale methods (\pyramidalflow, \edifyimage, and ours), where the default time segment is $T=0.5$. We use three levels for cases where resolutions are higher than $256 \times 256$, where time segments hyperparameters are $T_2=0.33$ and $T_1=0.67$. We use the linear scheduler instead of GVP unless explicitly mentioned. For ImageNet $256 \times 256$, we use a constant lr of $1\times10^{-4}$, a global batch size of $256$, and an EMA decay of $0.9999$. We train $600K$ steps for both the B/2 model and the XL/2 model. We further train the B/2 model until $7M$ steps following LFM~\citep{LFM}. For CelebA-HQ, we adopt a CosineAnnealingLR~\citep{CosineAnnealingLR} scheduler with an initial learning rate of $2 \times 10^{-4}$ and a final learning rate of $1 \times 10^{-6}$. Training durations vary across resolutions, as higher resolutions typically require more iterations to converge. A complete list of training hyperparameters can be found in~\Cref{sec-training-details}. All models are trained on a computing node with 8 NVIDIA H200 GPUs.
% \vspace{-4mm}
\subsection{Experimental Results}
% \vspace{-3mm}
\paragraph{Unconditional generation on CelebA-HQ.}
We evaluate our method on the CelebA-HQ dataset at resolutions of $256\times256$, $512\times512$, and $1024\times1024$, comparing against single-scale and multi-scale flow matching approaches. As shown in~\Cref{tab-celebahq-results}, our method achieves an FID of $3.53$ at $256\times256$, outperforming LFM ($5.26$) and Pyramidal Flow ($11.20$). The performance gap becomes even more pronounced at higher resolutions, where our method achieves FID scores of $4.04$ and $5.51$ at $512\times512$ and $1024\times1024$, respectively, compared to $6.35$ and $8.12$ for LFM. In particular, our method not only produces higher quality samples, but also requires fewer function evaluations (NFE) and less inference time compared to flow matching baselines (\lfm~\citep{LFM}, \pyramidalflow~\citep{PyramidFlow}, and~\edifyimage~\citep{EdifyImage}), demonstrating the efficiency of our multi-scale architecture in preserving image fidelity while reducing computational requirements. While Relay Diffusion obtains a competitive FID of $3.15$ at $256\times256$, it relies on image-space modeling and a U-Net backbone, leading to significantly higher inference cost (1221 GFLOPs vs.\ 16.5 GFLOPs for ours). This indicates a fundamentally different computational regime, as our latent-space DiT architecture provides a far more favorable tradeoff between image fidelity and efficiency. Furthermore, Relay Diffusion is strictly formulated as a two-stage framework and, to our knowledge, has not been demonstrated beyond $256\times256$ resolutions, whereas our Laplacian multi-scale formulation scales effectively to $512\times512$ and $1024\times1024$ while retaining efficiency gains. \Cref{fig-celeba-res} shows qualitative results of our method across different resolutions, highlighting high-fidelity detail preservation and demonstrating that our multi-scale formulation is effective even at megapixel resolution.

% ablation

% \vspace{-3mm}
\paragraph{Ablation studies.} 
Our comprehensive ablation studies on CelebA-HQ ($256 \times 256$ resolution) demonstrate several key insights for \modelname. As shown in \Cref{subtab-vae}, the choice of VAE architecture significantly impacts performance, with EQVAE providing substantial benefits to our approach while proving detrimental to the baseline LFM. The proposed mixture-of-transformers (MoT) design (\Cref{subtab-mot}) achieves both computational efficiency and improved generation quality compared to separate models. Furthermore, we find causal masking is optimal among attention strategies (\Cref{subtab-causal}). We then ablate the choice of critical time points. For two-scale experiments, we observe a critical time point of $T=0.5$ (\Cref{subtab:coupling}) strikes the ideal balance between model expressivity and complexity. Here we want to note that in two scales there is only one critical time point $T$. As summarized in~\Cref{subtab-noise} and further discussed in~\Cref{app:linear-schedules}, the linear pair $\sigma_t^{(k)} = 1 - t$ and $\alpha_t^{(k)} = \frac{t - T_{k+1}}{1 - T_{k+1}}$ achieves the strongest performance, outperforming quadratic and cubic decay as well as the GVP formulation. \textcolor{black}{As shown in~\Cref{subtab-num_scales}, two scales yield the best performance at 256×256. This is due to the latent resolution being only $32\times32$; adding a third scale would require an $8\times8$ stage, which we find too small to provide reliable semantic guidance during early flow integration. To further validate this resolution-dependent behavior, we extend this ablation to 512$^2$ and 1024$^2$ generation. As shown in Appendix~\ref{app:num-scales-higher-res}, larger latent grids (64$\times$64 and 128$\times$128) benefit from a more-scale hierarchy, confirming that higher resolutions may require more number of scales of~\modelname.
A similar trend is reported in the concurrent work PixelFlow~\citep{pixelflow}, which observes degraded performance when the initial kickoff sequence becomes too short, although their model operates in pixel space rather than latent space. To validate temporal segmentation beyond the two-scale case, we additionally ablate three-scale configurations and show that a balanced time allocation across scales yields the best results; see Appendix~\ref{app:temporal-segmentation} for FID comparisons at $1024 \times 1024$.}
We further study the effect of the noise and coupling schedules used in our flow parameterization. Specifically, we compare the linear transport and noise decays with alternative polynomial and GVP-based schedules. These results support our choice of simple scale-wise linear schedules for stable multi-scale flow evolution. The performance advantage of latent space modeling instead of the image space counterpart is particularly noteworthy (\Cref{subtab:nfe}), where our approach substantially outperforms both image-space implementations and the LFM baseline in two domains. \textcolor{black}{These findings, combined with our optimized learning rate decay with a final LR of $1\times 10^{-6}$ (\Cref{subtab:optimization}), collectively contribute to our full model.}

\begin{wrapfigure}{r}{0.55\textwidth}
  \centering
\includegraphics[width=\linewidth]{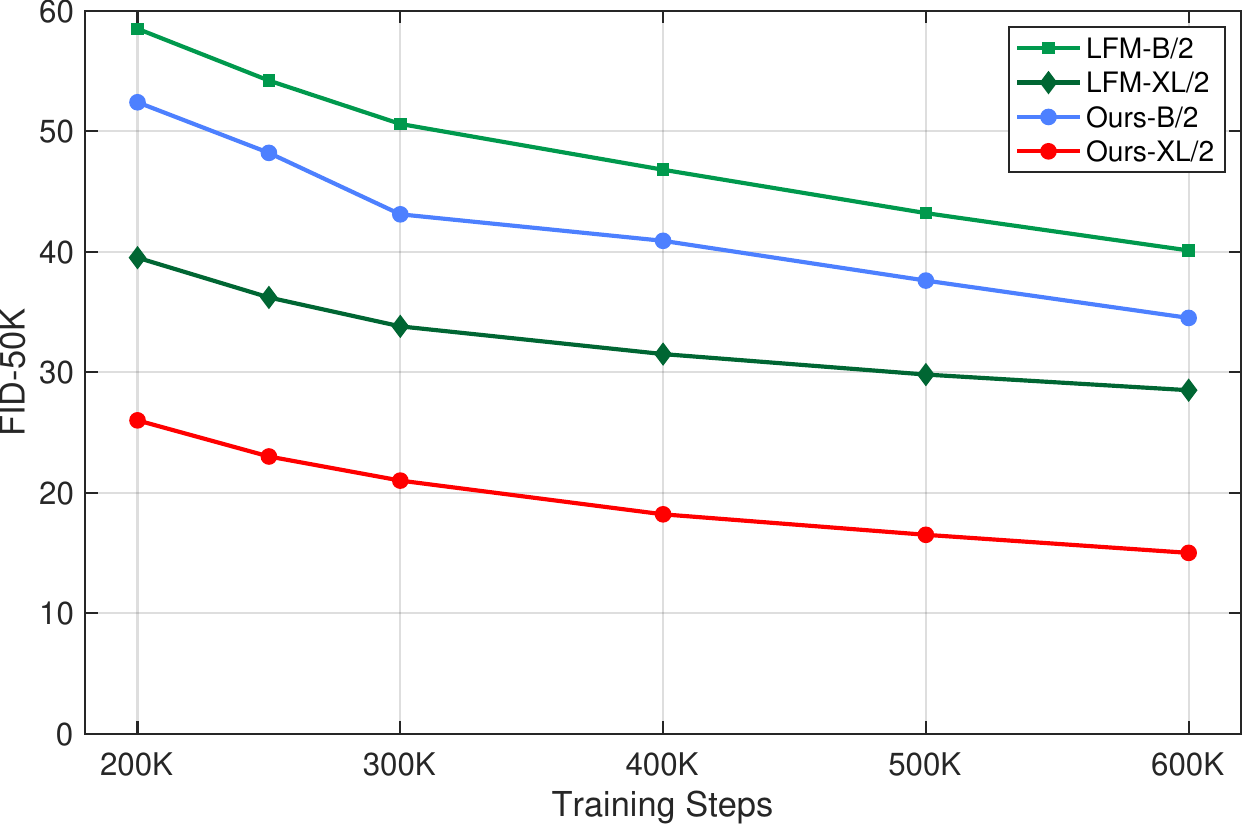}
\vspace{-3mm}
\caption{FID-50K on ImageNet (256$\times$256) across training iterations comparing LFM~\citep{LFM} with ours using two backbones (B/2 and XL/2).}
\label{fig-imagenet-training-curves}
\vspace{-2mm}
\end{wrapfigure}

\vspace{-1mm}
\paragraph{Class-conditional generation on ImageNet.} As illustrated in \Cref{fig-imagenet-training-curves}, our method consistently shows superior performance compared to LFM~\citep{LFM} for both B/2 and XL/2 models. We present numerical results in~\Cref{tab-imagenet-results} and qualitative results in~\Cref{fig-imagenet-sampling-res}. Our approach achieves the best FID score of 36.50 with the B/2 backbone and 14.38 with the XL/2 backbone at $600$K training steps, outperforming all baseline methods including DiT~\citep{DiT}, LFM~\citep{LFM}, and Pyramidal Flow~\citep{PyramidFlow}. Notably, our method works under fewer NFE (80 vs. 89-250) and demonstrates improved computational efficiency with lower inference time (1.25s vs. 1.70-1.87s for B/2) and reduced GFLOPs (4.9 vs. 5.6-5.7 for B/2). When trained for $7$M steps with classifier-free guidance (CFG=1.5), our approach achieves an FID score of 4.12, surpassing LFM's 4.46 while being more efficient. 
% \vspace{-3mm}
\section{Conclusion}
% \vspace{-1mm}
In this paper, we presented \modelname, a novel framework that advances generative modeling by integrating parallel multi-scale generation within the flow matching paradigm. Our approach decomposes images into Laplacian pyramid residuals and processes them simultaneously through a specialized MOT architecture with causal attention mechanisms that enforce hierarchical information flow. Extensive experimental results on CelebA-HQ and ImageNet datasets demonstrate that our method consistently achieves superior FID scores than prior flow matching works while requiring fewer function evaluations and reduced computational resources during sampling. Performance advantages are particularly pronounced at higher resolutions (up to 1024×1024), demonstrating its superior scalability for complex, detail-rich generation tasks even with limited training resources. \textcolor{black}{An interesting future direction is to integrate advanced training accelerators such as REPA~\citep{yu2024repa}, which has demonstrated significantly faster convergence for flow-based generative models. We expect that combining training accelerators with the multi-scale Laplacian formulation could further reduce training cost.}
\clearpage
\newpage

\section{Acknowledgments}
The authors are supported in part by grant NSF 2409016.


\bibliography{ICLR2026/example_paper}

\begin{thebibliography}{35}
\providecommand{\natexlab}[1]{#1}
\providecommand{\url}[1]{\texttt{#1}}
\expandafter\ifx\csname urlstyle\endcsname\relax
  \providecommand{\doi}[1]{doi: #1}\else
  \providecommand{\doi}{doi: \begingroup \urlstyle{rm}\Url}\fi

\bibitem[Albergo and Vanden-Eijnden(2022)]{albergo2022building}
Michael~S Albergo and Eric Vanden-Eijnden.
\newblock Building normalizing flows with stochastic interpolants.
\newblock \emph{arXiv preprint arXiv:2209.15571}, 2022.

\bibitem[Atzmon et~al.(2024)Atzmon, Bala, Balaji, Cai, Cui, Fan, Ge, Gururani, Huffman, Isaac, Jannaty, Karras, Lam, Lewis, Licata, Lin, Liu, Ma, Mallya, Martino-Tarr, Mendez, Nah, Pruett, Reda, Song, Wang, Wei, Zeng, Zeng, and Zhang]{EdifyImage}
Yuval Atzmon, Maciej Bala, Yogesh Balaji, Tiffany Cai, Yin Cui, Jiaojiao Fan, Yunhao Ge, Siddharth Gururani, Jacob Huffman, Ronald Isaac, Pooya Jannaty, Tero Karras, Grace Lam, J.~P. Lewis, Aaron Licata, Yen-Chen Lin, Ming-Yu Liu, Qianli Ma, Arun Mallya, Ashlee Martino-Tarr, Doug Mendez, Seungjun Nah, Chris Pruett, Fitsum Reda, Jiaming Song, Ting-Chun Wang, Fangyin Wei, Xiaohui Zeng, Yu~Zeng, and Qinsheng Zhang.
\newblock Edify image: High-quality image generation with pixel space laplacian diffusion models, 2024.
\newblock URL \url{https://arxiv.org/abs/2411.07126}.

\bibitem[Batzolis et~al.(2022)Batzolis, Stanczuk, Schönlieb, and Etmann]{NonUniformDiffusion}
Georgios Batzolis, Jan Stanczuk, Carola-Bibiane Schönlieb, and Christian Etmann.
\newblock Non-uniform diffusion models, 2022.
\newblock URL \url{https://arxiv.org/abs/2207.09786}.

\bibitem[Chen et~al.(2018)Chen, Rubanova, Bettencourt, and Duvenaud]{chen2018neuralode}
Ricky T.~Q. Chen, Yulia Rubanova, Jesse Bettencourt, and David Duvenaud.
\newblock Neural ordinary differential equations.
\newblock \emph{Advances in Neural Information Processing Systems}, 2018.

\bibitem[Chen et~al.(2025)Chen, Ge, Zhang, Sun, and Luo]{pixelflow}
Shoufa Chen, Chongjian Ge, Shilong Zhang, Peize Sun, and Ping Luo.
\newblock Pixelflow: Pixel-space generative models with flow, 2025.
\newblock URL \url{https://arxiv.org/abs/2504.07963}.

\bibitem[Dao et~al.(2023)Dao, Phung, Nguyen, and Tran]{LFM}
Quan Dao, Hao Phung, Binh Nguyen, and Anh Tran.
\newblock Flow matching in latent space, 2023.
\newblock URL \url{https://arxiv.org/abs/2307.08698}.

\bibitem[Deng et~al.(2009)Deng, Dong, Socher, Li, Li, and Fei-Fei]{imageNet}
Jia Deng, Wei Dong, Richard Socher, Li-Jia Li, Kai Li, and Li~Fei-Fei.
\newblock Imagenet: A large-scale hierarchical image database.
\newblock In \emph{2009 IEEE Conference on Computer Vision and Pattern Recognition}, pages 248--255, 2009.
\newblock \doi{10.1109/CVPR.2009.5206848}.

\bibitem[Dhariwal and Nichol(2021)]{dhariwal2021diffusion}
Prafulla Dhariwal and Alexander~Quinn Nichol.
\newblock Diffusion models beat {GAN}s on image synthesis.
\newblock In A.~Beygelzimer, Y.~Dauphin, P.~Liang, and J.~Wortman Vaughan, editors, \emph{Advances in Neural Information Processing Systems}, 2021.
\newblock URL \url{https://openreview.net/forum?id=AAWuCvzaVt}.

\bibitem[Dosovitskiy et~al.(2020)Dosovitskiy, Beyer, Kolesnikov, Weissenborn, Zhai, Unterthiner, Dehghani, Minderer, Heigold, Gelly, et~al.]{ViT}
Alexey Dosovitskiy, Lucas Beyer, Alexander Kolesnikov, Dirk Weissenborn, Xiaohua Zhai, Thomas Unterthiner, Mostafa Dehghani, Matthias Minderer, Georg Heigold, Sylvain Gelly, et~al.
\newblock An image is worth 16x16 words: Transformers for image recognition at scale.
\newblock \emph{arXiv preprint arXiv:2010.11929}, 2020.

\bibitem[Esser et~al.(2024)Esser, Kulal, Blattmann, Entezari, M{\"u}ller, Saini, Levi, Lorenz, Sauer, Boesel, et~al.]{esser2024scaling}
Patrick Esser, Sumith Kulal, Andreas Blattmann, Rahim Entezari, Jonas M{\"u}ller, Harry Saini, Yam Levi, Dominik Lorenz, Axel Sauer, Frederic Boesel, et~al.
\newblock Scaling rectified flow transformers for high-resolution image synthesis.
\newblock In \emph{Forty-first international conference on machine learning}, 2024.

\bibitem[Fedus et~al.(2022)Fedus, Zoph, and Shazeer]{fedus2022switch}
William Fedus, Barret Zoph, and Noam Shazeer.
\newblock Switch transformers: Scaling to trillion parameter models with simple and efficient sparsity.
\newblock \emph{Journal of Machine Learning Research}, 23\penalty0 (120):\penalty0 1--39, 2022.

\bibitem[{Google DeepMind}(2025)]{deepmind2025veo2}
{Google DeepMind}.
\newblock Veo 2.
\newblock \url{https://deepmind.google/technologies/veo/veo-2/}, 2025.
\newblock Accessed: 2025-05-09.

\bibitem[Heusel et~al.(2017)Heusel, Ramsauer, Unterthiner, Nessler, and Hochreiter]{heusel2017gansFID}
Martin Heusel, Hubert Ramsauer, Thomas Unterthiner, Bernhard Nessler, and Sepp Hochreiter.
\newblock Gans trained by a two time-scale update rule converge to a local nash equilibrium.
\newblock \emph{Advances in neural information processing systems}, 30, 2017.

\bibitem[Ho and Salimans(2022)]{ho2022classifier}
Jonathan Ho and Tim Salimans.
\newblock Classifier-free diffusion guidance.
\newblock \emph{arXiv preprint arXiv:2207.12598}, 2022.

\bibitem[Ho et~al.(2020)Ho, Jain, and Abbeel]{ddpm}
Jonathan Ho, Ajay Jain, and Pieter Abbeel.
\newblock Denoising diffusion probabilistic models.
\newblock \emph{Advances in Neural Information Processing Systems}, 33:\penalty0 6840--6851, 2020.

\bibitem[Ho et~al.(2022)Ho, Saharia, Chan, Fleet, Norouzi, and Salimans]{ho2022cascaded}
Jonathan Ho, Chitwan Saharia, William Chan, David~J Fleet, Mohammad Norouzi, and Tim Salimans.
\newblock Cascaded diffusion models for high fidelity image generation.
\newblock \emph{Journal of Machine Learning Research}, 23\penalty0 (47):\penalty0 1--33, 2022.

\bibitem[Jin et~al.(2025)Jin, Sun, Li, Xu, Xu, Jiang, Zhuang, Huang, Song, MU, and Lin]{PyramidFlow}
Yang Jin, Zhicheng Sun, Ningyuan Li, Kun Xu, Kun Xu, Hao Jiang, Nan Zhuang, Quzhe Huang, Yang Song, Yadong MU, and Zhouchen Lin.
\newblock Pyramidal flow matching for efficient video generative modeling.
\newblock In \emph{The Thirteenth International Conference on Learning Representations}, 2025.
\newblock URL \url{https://openreview.net/forum?id=66NzcRQuOq}.

\bibitem[Karras et~al.(2018)Karras, Aila, Laine, and Lehtinen]{ProgressiveGan}
Tero Karras, Timo Aila, Samuli Laine, and Jaakko Lehtinen.
\newblock Progressive growing of {GAN}s for improved quality, stability, and variation.
\newblock In \emph{International Conference on Learning Representations}, 2018.
\newblock URL \url{https://openreview.net/forum?id=Hk99zCeAb}.

\bibitem[Kouzelis et~al.(2025)Kouzelis, Kakogeorgiou, Gidaris, and Komodakis]{EQVAE}
Theodoros Kouzelis, Ioannis Kakogeorgiou, Spyros Gidaris, and Nikos Komodakis.
\newblock Eq-vae: Equivariance regularized latent space for improved generative image modeling, 2025.
\newblock URL \url{https://arxiv.org/abs/2502.09509}.

\bibitem[Lai et~al.(2017)Lai, Huang, Ahuja, and Yang]{LapGAN}
Wei-Sheng Lai, Jia-Bin Huang, Narendra Ahuja, and Ming-Hsuan Yang.
\newblock Deep laplacian pyramid networks for fast and accurate super-resolution.
\newblock In \emph{Proceedings of the IEEE conference on computer vision and pattern recognition}, pages 624--632, 2017.

\bibitem[Liang et~al.(2024)Liang, Yu, Luo, Iyer, Dong, Zhou, Ghosh, Lewis, Yih, Zettlemoyer, et~al.]{MoT}
Weixin Liang, Lili Yu, Liang Luo, Srinivasan Iyer, Ning Dong, Chunting Zhou, Gargi Ghosh, Mike Lewis, Wen-tau Yih, Luke Zettlemoyer, et~al.
\newblock Mixture-of-transformers: A sparse and scalable architecture for multi-modal foundation models.
\newblock \emph{arXiv preprint arXiv:2411.04996}, 2024.

\bibitem[Lipman et~al.(2023)Lipman, Chen, Ben-Hamu, Nickel, and Le]{FlowMatchingGenerativeModeling}
Yaron Lipman, Ricky T.~Q. Chen, Heli Ben-Hamu, Maximilian Nickel, and Matthew Le.
\newblock Flow matching for generative modeling.
\newblock In \emph{The Eleventh International Conference on Learning Representations}, 2023.
\newblock URL \url{https://openreview.net/forum?id=PqvMRDCJT9t}.

\bibitem[Liu et~al.(2022)Liu, Gong, and Liu]{liu2022flow}
Xingchao Liu, Chengyue Gong, and Qiang Liu.
\newblock Flow straight and fast: Learning to generate and transfer data with rectified flow.
\newblock \emph{arXiv preprint arXiv:2209.03003}, 2022.

\bibitem[Loshchilov and Hutter(2017)]{CosineAnnealingLR}
Ilya Loshchilov and Frank Hutter.
\newblock {SGDR}: Stochastic gradient descent with warm restarts.
\newblock In \emph{International Conference on Learning Representations}, 2017.
\newblock URL \url{https://openreview.net/forum?id=Skq89Scxx}.

\bibitem[Ma et~al.(2024)Ma, Goldstein, Albergo, Boffi, Vanden-Eijnden, and Xie]{ma2024sit}
Nanye Ma, Mark Goldstein, Michael~S Albergo, Nicholas~M Boffi, Eric Vanden-Eijnden, and Saining Xie.
\newblock Sit: Exploring flow and diffusion-based generative models with scalable interpolant transformers.
\newblock In \emph{European Conference on Computer Vision}, pages 23--40. Springer, 2024.

\bibitem[Peebles and Xie(2023)]{DiT}
William Peebles and Saining Xie.
\newblock Scalable diffusion models with transformers.
\newblock In \emph{Proceedings of the IEEE/CVF International Conference on Computer Vision}, pages 4195--4205, 2023.

\bibitem[Rombach et~al.(2022)Rombach, Blattmann, Lorenz, Esser, and Ommer]{LDM}
Robin Rombach, Andreas Blattmann, Dominik Lorenz, Patrick Esser, and Bj{\"o}rn Ommer.
\newblock High-resolution image synthesis with latent diffusion models.
\newblock In \emph{Proceedings of the IEEE/CVF conference on computer vision and pattern recognition}, pages 10684--10695, 2022.

\bibitem[Saharia et~al.(2022)Saharia, Chan, Saxena, Li, Whang, Denton, Ghasemipour, Gontijo~Lopes, Karagol~Ayan, Salimans, et~al.]{imagen}
Chitwan Saharia, William Chan, Saurabh Saxena, Lala Li, Jay Whang, Emily~L Denton, Kamyar Ghasemipour, Raphael Gontijo~Lopes, Burcu Karagol~Ayan, Tim Salimans, et~al.
\newblock Photorealistic text-to-image diffusion models with deep language understanding.
\newblock \emph{Advances in neural information processing systems}, 35:\penalty0 36479--36494, 2022.

\bibitem[Shazeer et~al.(2017)Shazeer, Mirhoseini, Maziarz, Davis, Le, Hinton, and Dean]{shazeer2017outrageously}
Noam Shazeer, Azalia Mirhoseini, Krzysztof Maziarz, Andy Davis, Quoc Le, Geoffrey Hinton, and Jeff Dean.
\newblock Outrageously large neural networks: The sparsely-gated mixture-of-experts layer.
\newblock In \emph{International Conference on Learning Representations (ICLR)}, 2017.

\bibitem[Sun et~al.(2024)Sun, Jiang, Chen, Zhang, Peng, Luo, and Yuan]{LlamaGen}
Peize Sun, Yi~Jiang, Shoufa Chen, Shilong Zhang, Bingyue Peng, Ping Luo, and Zehuan Yuan.
\newblock Autoregressive model beats diffusion: Llama for scalable image generation.
\newblock \emph{arXiv preprint arXiv:2406.06525}, 2024.

\bibitem[Teng et~al.(2024)Teng, Zheng, Ding, Hong, Wangni, Yang, and Tang]{relaydiffusion}
Jiayan Teng, Wendi Zheng, Ming Ding, Wenyi Hong, Jianqiao Wangni, Zhuoyi Yang, and Jie Tang.
\newblock Relay diffusion: Unifying diffusion process across resolutions for image synthesis.
\newblock In \emph{The Twelfth International Conference on Learning Representations}, 2024.
\newblock URL \url{https://openreview.net/forum?id=qTlcbLSm4p}.

\bibitem[Tian et~al.(2024)Tian, Jiang, Yuan, PENG, and Wang]{VAR}
Keyu Tian, Yi~Jiang, Zehuan Yuan, BINGYUE PENG, and Liwei Wang.
\newblock Visual autoregressive modeling: Scalable image generation via next-scale prediction.
\newblock In \emph{The Thirty-eighth Annual Conference on Neural Information Processing Systems}, 2024.
\newblock URL \url{https://openreview.net/forum?id=gojL67CfS8}.

\bibitem[Vaswani et~al.(2017)Vaswani, Shazeer, Parmar, Uszkoreit, Jones, Gomez, Kaiser, and Polosukhin]{transformer}
Ashish Vaswani, Noam Shazeer, Niki Parmar, Jakob Uszkoreit, Llion Jones, Aidan~N Gomez, \L~ukasz Kaiser, and Illia Polosukhin.
\newblock Attention is all you need.
\newblock In I.~Guyon, U.~Von Luxburg, S.~Bengio, H.~Wallach, R.~Fergus, S.~Vishwanathan, and R.~Garnett, editors, \emph{Advances in Neural Information Processing Systems}, volume~30. Curran Associates, Inc., 2017.
\newblock URL \url{https://proceedings.neurips.cc/paper_files/paper/2017/file/3f5ee243547dee91fbd053c1c4a845aa-Paper.pdf}.

\bibitem[Yu et~al.(2025)Yu, Kwak, Jang, Jeong, Huang, Shin, and Xie]{yu2024repa}
Sihyun Yu, Sangkyung Kwak, Huiwon Jang, Jongheon Jeong, Jonathan Huang, Jinwoo Shin, and Saining Xie.
\newblock Representation alignment for generation: Training diffusion transformers is easier than you think.
\newblock In \emph{International Conference on Learning Representations}, 2025.

\bibitem[Zhang et~al.(2023)Zhang, Wang, Zhang, Zhao, Yuan, Qin, Wang, Zhao, and Zhou]{cascadedvideo}
Shiwei Zhang, Jiayu Wang, Yingya Zhang, Kang Zhao, Hangjie Yuan, Zhiwu Qin, Xiang Wang, Deli Zhao, and Jingren Zhou.
\newblock I2vgen-xl: High-quality image-to-video synthesis via cascaded diffusion models.
\newblock \emph{arXiv preprint arXiv:2311.04145}, 2023.

\end{thebibliography}
\bibliographystyle{plainnat}
\clearpage
\newpage

\appendix
\section*{Appendix}
\section{Temporal Segmentation Threshold Validation}
\label{app:temporal-segmentation}

As discussed in the main paper, our progressive multi-scale diffusion training strategy allocates timesteps across spatial scales using temporal segmentation thresholds. While~\Cref{tab:ablation}(d) presents an ablation on a single segmentation point \(T\) in a two-scale setting, we provide additional justification for the selected three-scale thresholds. To address this concern, we provide extended ablation experiments on the temporal segmentation thresholds \(T_1\) and \(T_2\) for a three-scale configuration at a resolution of \(1024 \times 1024\). Except for the threshold values, all other training configurations are identical to~\Cref{tab:ablation}.

\textcolor{black}{Results are presented in~\Cref{tab:three-scale-tseg}. These results empirically validate our threshold choice in the main paper, demonstrating that a balanced temporal allocation across scales (\(T_1 = 0.67, T_2 = 0.33\)) yields the best generation quality. This trend is consistent with our two-scale observations, where overly early or late transitions led to degraded FID.}

\textcolor{black}{We hypothesize that allocating sufficient diffusion steps to the coarse-scale stage stabilizes global structures before introducing high-frequency refinements in later stages. In contrast, excessively delayed transitions (e.g., \(T_1=0.90\)) limit the optimization of high-resolution layers, while overly early transitions (e.g., \(T_2 = 0.10\)) destabilize global consistency and lead to artifacts.}

\textcolor{black}{A complete grid search for optimal segmentation is computationally prohibitive at high resolutions; however, the reported results provide strong empirical justification for our selected thresholds.}

\textcolor{black}{\textbf{Practical Recommendation.} Based on this analysis, we recommend the following default heuristic for future multi-scale diffusion training:
\begin{equation}
T_1 \approx \frac{2}{3}, \qquad T_2 \approx \frac{1}{3},
\end{equation}
which we found to consistently improve stability and generation quality.}

\section{\textcolor{black}{Additional Analysis: Number of Scales at Higher Resolutions}}
\label{app:num-scales-higher-res}
\textcolor{black}{
In the main paper, we show that at $256^2$ resolution, a two-scale Laplacian hierarchy achieves the best performance. Here, we extend this analysis to higher resolutions.}

\textcolor{black}{As the resolution increases, the VAE latent grids become larger:
\[
256^2 \rightarrow 32\times 32, \quad
512^2 \rightarrow 64\times 64, \quad
1024^2 \rightarrow 128\times 128.
\]
We present a further study at larger scales. Other unmentioned setups are the same as~\Cref{subtab-num_scales}.}

\begin{table}[h]
\centering
\caption{\textcolor{black}{\textbf{FID vs. number of Laplacian scales} across 256–1024 resolutions on CelebA-HQ.}}
\textcolor{black}{\begin{tabular}{c|c|c|c|c}
\toprule
Resolution & Latent size & \# Scales & Decomposition & FID $\downarrow$ \\
\midrule
256$\times$256  & 32$\times$32  & 2 & 32 $\rightarrow$ 16 & \textbf{3.53} \\
256$\times$256  & 32$\times$32  & 3 & 32 $\rightarrow$ 16 $\rightarrow$ 8 & 3.59 \\
256$\times$256  & 32$\times$32  & 4 & 32 $\rightarrow$ 16 $\rightarrow$ 8 $\rightarrow$ 4 & 5.12 \\
\midrule
512$\times$512  & 64$\times$64  & 2 & 64 $\rightarrow$ 32 & 5.45 \\
512$\times$512  & 64$\times$64  & 3 & 64 $\rightarrow$ 32 $\rightarrow$ 16 & \textbf{4.04} \\
512$\times$512  & 64$\times$64  & 4 & 64 $\rightarrow$ 32 $\rightarrow$ 16 $\rightarrow$ 8 & 4.12 \\
\midrule
1024$\times$1024 & 128$\times$128 & 2 & 128 $\rightarrow$ 64 & 6.62 \\
1024$\times$1024 & 128$\times$128 & 3 & 128 $\rightarrow$ 64 $\rightarrow$ 32 & 5.51 \\
1024$\times$1024 & 128$\times$128 & 4 & 128 $\rightarrow$ 64 $\rightarrow$ 32 $\rightarrow$ 16 & \textbf{5.45} \\
\bottomrule
\end{tabular}}
\vspace{-3mm}
\end{table}

\textcolor{black}{
At low resolution, additional Laplacian levels produce coarse grids too small to stabilize flow trajectories, whereas at higher resolution, three-scale Laplacian modeling delivers consistent improvements.
These results are consistent with observations in concurrent PixelFlow~\citep{pixelflow}, which reports degraded performance when early-stage spatial support becomes too small, despite operating in pixel (not latent) space.}

\section{\textcolor{black}{Time-Weighted Complexity Analysis of Multi-scale MoT}}
\label{app:mot-complexity}

\textcolor{black}{
We provide a formal complexity analysis of the proposed Mixture-of-Transformers (MoT) architecture, taking into account that different Laplacian scales are active over different segments of the ODE time domain. Although global attention is applied over the concatenated token sequence of all active scales at each time segment, the progressively shortened active ranges of higher-resolution scales reduce the time-averaged number of tokens processed during sampling.
}

\textcolor{black}{
Under the standard quadratic attention cost used in transformer analysis~\citep{transformer}, the per-layer cost scales as $N^2d$, where $N$ is the number of latent tokens and $d$ is the embedding dimension; full self-attention scales quadratically with sequence length.}

\textcolor{black}{
Let the finest latent resolution be $H \times W$ with downsampling factor $s$, so the largest-scale latent has $N = \frac{HW}{s^2}$ tokens. Following the Laplacian decomposition in~\Cref{eq-laplacian-decomposition-combine}, the \emph{per-scale} token counts at the three spatial resolutions are
$N^{(0)} = N$, $N^{(1)} = N/4$, and $N^{(2)} = N/16$.
During sampling, the ODE is solved in three segments:
from $t \in [0, T_2)$ only the coarsest scale is active,
from $t \in [T_2, T_1)$ the coarsest and mid scales are active,
and from $t \in [T_1, 1]$ all three scales are active.
Thus, the total number of tokens processed in each segment is
}
\textcolor{black}{
\begin{equation}
N_2 = N^{(2)} = \tfrac{N}{16}, \qquad
N_1 = N^{(1)} + N^{(2)} = \tfrac{5N}{16}, \qquad
N_0 = N^{(0)} + N^{(1)} + N^{(2)} = \tfrac{21N}{16}.
\end{equation}
}
\textcolor{black}{
Let the segment lengths be
$\Delta t_2 = T_2 - 0$, $\Delta t_1 = T_1 - T_2$, and $\Delta t_0 = 1 - T_1$.
With our default choice $T_2 \approx \tfrac{1}{3}$ and $T_1 \approx \tfrac{2}{3}$ (Sec.~\ref{sec-multi-scale-sampling-process}), we have $\Delta t_2 \approx \Delta t_1 \approx \Delta t_0 \approx \tfrac{1}{3}$.
}
\begin{table}[t]
\centering
\textcolor{black}{
\begin{tabular}{c|c|c|c}
\toprule
Resolution & \(T_1\) & \(T_2\) & FID$\downarrow$ \\
\midrule
1024$\times$1024 & 0.67 & 0.33 & \textbf{5.51} \\
1024$\times$1024 & 0.90 & 0.33 & 6.72 \\
1024$\times$1024 & 0.67 & 0.10 & 8.12 \\
\bottomrule
\end{tabular}}
\caption{\textcolor{black}{Three-scale temporal segmentation ablation at \(1024^2\) resolution. Balanced allocation (\(T_1=0.67, T_2=0.33\)) achieves the best FID, confirming the trend observed in the two-scale case.}}
\label{tab:three-scale-tseg}
\end{table}
\textcolor{black}{
Because attention cost in the $k$-th segment is proportional to $N_k^2 d$, the time-weighted attention complexity of one MoT layer over the whole ODE trajectory is
}
\textcolor{black}{
\begin{equation}
\text{Cost}_{\text{MoT}}
\;\propto\;
\Delta t_2 N_2^2d + \Delta t_1 N_1^2d + \Delta t_0 N_0^2d
\;\approx\;
\tfrac{1}{3}\!\left[
\Big(\tfrac{N}{16}\Big)^2
+
\Big(\tfrac{5N}{16}\Big)^2
+
\Big(\tfrac{21N}{16}\Big)^2
\right]d\approx 0.61\,N^2d.
\end{equation}
}
\textcolor{black}{
Therefore, under our default three-scale configuration the MoT layer enjoys a reduction factor of about $0.61$ in its attention term, i.e., roughly a $1.6\times$ lower attention cost than a single-scale DiT with the same finest latent size.
}

\textcolor{black}{
We emphasize that this is an idealized, time-weighted analysis focusing on the dominant quadratic attention term. Empirical GFLOPs and runtime may differ from this constant factor due to (i) the adaptive ODE solver (Dormand–Prince) producing different numbers of function evaluations across samples, (ii) kernel-level hardware efficiency, (iii) upsampling and residual fusion costs in Laplacian reconstruction, and (iv) framework-dependent attention kernel implementations. Nevertheless, the observed improvements in Tables~\ref{tab-celebahq-results},~\ref{tab:ablation}, and~\ref{tab-imagenet-results} are consistent with this theoretical reduction in effective attention cost.
}

\section{\textcolor{black}{Additional Discussion on Linear Transport and Noise Schedules}}
\label{app:linear-schedules}

\textcolor{black}{
\textbf{Clarification of schedule formulation.}
Both the noise decay schedule
$\sigma_t^{(k)} = 1 - t$
and the transport (coupling) schedule
$\alpha_t^{(k)} = \frac{t - T_{k+1}}{1 - T_{k+1}}$
are \textbf{linear functions of $t$}. In particular,
\begin{equation}
\alpha_t^{(k)}
= \frac{t - T_{k+1}}{1 - T_{k+1}}
= \frac{1}{1 - T_{k+1}}\,t
  - \frac{T_{k+1}}{1 - T_{k+1}},
\end{equation}
which is a linear mapping that increases from $0$ to $1$ as $t$ moves from $T_{k+1}$ to $1$. The two linear schedules play distinct roles: $\sigma_t^{(k)}$ is globally shared across all scales, while $\alpha_t^{(k)}$ provides a shift--rescale transformation that activates Laplacian levels gradually after their transition threshold $T_{k+1}$. This design offers a simple and stable mechanism for connecting consecutive Laplacian scales \emph{without} requiring re-noising steps used in related multi-scale flow models such as~\cite{PyramidFlow}.
}

\vspace{3pt}

\textcolor{black}{
\textbf{Ablation on alternative schedule choices.}
To support this design choice, we further evaluate several alternative combinations of $\sigma_t^{(k)}$ and $\alpha_t^{(k)}$ at the $256 \times 256$ resolution. This experiment extends the analysis in~\Cref{subtab-noise} while keeping all other settings identical to the default configuration in~\Cref{tab:ablation}.
}

\begin{table}[h]
\centering
\textcolor{black}{
\begin{tabular}{l|l|c}
\toprule
\textbf{Noise schedule $\sigma_t^{(k)}$} &
\textbf{Coupling schedule $\alpha_t^{(k)}$} &
\textbf{FID $\downarrow$ (256$^2$)}
\\ \midrule
$1 - t$ (Linear) &
$\frac{t - T_{k+1}}{1 - T_{k+1}}$ (Linear) &
\textbf{3.53}
\\
$\cos\!\left(\tfrac{\pi}{2} t\right)$ (GVP) &
$\sin\!\left(\tfrac{\pi (t - T_{k+1})}{2(1 - T_{k+1})}\right)$ (GVP) &
4.10
\\
$(1 - t)^2$ (Quadratic) &
$\frac{t - T_{k+1}}{1 - T_{k+1}}$ (Linear) &
3.78
\\
$(1 - t)^3$ (Cubic) &
$\frac{t - T_{k+1}}{1 - T_{k+1}}$ (Linear) &
4.25
\\
\bottomrule
\end{tabular}
}
\vspace{-2mm}
\end{table}

\noindent
\textcolor{black}{
These results indicate that the simple linear schedule $\sigma_t^{(k)} = 1 - t$ achieves the best overall performance, while steeper decays such as $(1-t)^2$ and $(1-t)^3$ lead to noticeably worse sample quality. The GVP variant also performs competitively, but the linear pair $(\sigma_t^{(k)}, \alpha_t^{(k)})$ is both \emph{simpler} and \emph{more stable} across a wide range of settings, reinforcing the choice of linear schedules as the default configuration in our framework.
}

\section{\edifyimage Formulation Details}
\label{sec-EdifyImage-appendix}
The noising process is described as a weighted interpolation between the clean image $\bx_1$ and the standard Gaussian noise $\bx_0^{(0)} \sim \cN(0, I^{(0)})$ (note this is not a residual) corresponding to the largest scale, where we get a noisy state $\bx_t^{(k)}$ for scale $k=0,1,2$ via:
\begin{equation}
\bx_t^{(k)} = (1-t)\down(\bx_0^{(0)}, 2^{k}) + \mu^{(k)}(\bx_1, t),
\label{eq-edity-image-xt}
\end{equation}
where we use a factor $2^{k}$ for better compatibility of sampling across multiple scales, and we provide more discussions on this in the supplementary.

Based on the noisy target $\bx_t^{(k)}$, we construct the following ground-true velocity:
\begin{equation}
\begin{aligned}
\bu_t^{(k)}(\bx_t^{(k)}|\bx_1) &\triangleq \frac{d\bx_t^{(k)}}{dt}\\& =-\down(\bx_0^{(0)}, 2^{k}) + \frac{d\mu^{(k)}(\bx_1, t)}{dt},
\end{aligned}
\label{eq-EdifyImage-gt-velocity}
\end{equation}
% \begin{equation}
% \bu_t^{(k)}(\bx_t^{(k)}|\bx_1) \triangleq \frac{d\bx_t^{(k)}}{dt} = 
% -2^{(k)}\epsilon^{(k)} + \mu^{(k)}(\bx_1, t) + t\frac{d\mu^{(k)}(\bx_1, t)}{dt},
% \end{equation}
where the last term is crucial because the data mean $\mu^{(k)}(\bx_1, t)$ changes through timestep $t$. Based on this velocity, the model is trained through the following flow-matching target~\citep{FlowMatchingGenerativeModeling}: 
\begin{equation}
\mE_{t\in\left[T_{k+1}, 1\right], q(\bx_1), p_t(\bx_t^{(k)}|\bx_1)}\left\|\bv_t^{(k)}(\bx_t^{(k)})- \bu_t^{(k)}(\bx_t^{(k)}|\bx_1)\right\|^2,
\label{eq-EdifyImage-loss}
\end{equation}
where $t$ is sampled from $[T_{k+1}, 1]$ (we define $T_3\triangleq0$, and $T_0\triangleq1$), $p_t(\bx_t^{(k)}|\bx_1)$ is the underlying conditional probability path towards $\bx_1$ determined by $\bu_t^{(k)}(\bx_t^{(k)}|\bx_1)$, and $\bv_t^{(k)}(\bx_t^{(k)})$ is the multi-scale machine learning model to model the flow. 

\section{\pyramidalflow}
\begin{algorithm}
   \caption{Pyramidal Flow Training}
   \label{pf-train-code}
    \begin{algorithmic}[1]
    \STATE \textbf{Input:} Dataset of images $D$, number of stages $K$, number of epochs $M$, list of $K$ start and end times $s_k$, $e_k$
    \STATE \textbf{Output:} Trained flow model $\bv(\bx, t)$.
    \STATE Initialize the model weights randomly
    \FOR{epoch$= 1$ to $M$}
        \FOR{clean data $\bx_1$ in $D$}
            \STATE Determine the stage k for $\bx_1$
            \STATE Sample a timestep $t \sim U(s_k, e
            _k)$
            \STATE Sample the start point $\bx_0 \sim \cN(0, I)$
            \STATE Down-sample noise $\bx_0^{(k)} = \down(\bx_0, 2^{k})$
            \STATE Compute start points as $\bx_s^{(k)} = s_k \cdot \up(\down(\bx_1, 2^{k+1})) + (1 - s_k) \cdot \bx_0^{(k)}$
            \STATE Compute endpoints as $\bx_e^{(k)} = e_k \cdot \down(\bx_1, 2^{k}) + (1 - e_k) \cdot \bx_0^{(k)}$
            \STATE Compute the midpoints as $\bx_t^{(k)} = \frac{e_k-t}{e_k-s_k} \cdot \bx_s^{(k)} + \frac{t-s_k}{e_k-s_k} \cdot \bx_e^{(k)}$
            \STATE Define the target as $\bu_t(\bx_t^{(k)}|\bx_1) = \frac{\bx_e^{(k)} - \bx_s^{(k)}}{e_k-s_k}$
            \STATE Calculate loss as $\operatorname{MSE}(\bu_t(\bx_t^{(k)}|\bx_1), \textbf{v}(\bx_t^{(k)}))$
            \STATE Run back-propagation, update parameters of $\bv$
        \ENDFOR
    \ENDFOR
    \STATE \textbf{Return:} $\bv$
    \end{algorithmic}
\end{algorithm}
\begin{algorithm}
   \caption{Pyramidal Flow Sampling}
   \label{pf-sample-code}
    \begin{algorithmic}[1]
    \STATE \textbf{Input:} Trained flow model $\bv$, number of stages $K$, list of $K$ start times $s_k$, total number of timesteps $N$.
    \STATE \textbf{Output:} Generated image.
    \STATE Sample the starting point $\hat{\bx}_0^{K-1} \sim \cN\left(0, I \cdot \frac{1}{4^{K-1}}\right)$
    \STATE Sample Gaussian noise at maximum resolution $\bx_0 \sim \cN(0, I)$
    \FOR{stage $k = K-1$ to $0$}
        \STATE $\hat{\bx}_{s_{k-1}}^{(k-1)} = \operatorname{ODEINT}(\bv_t^{(k)}, \{s_k, s_{k-1}\}, \hat{\bx}_{s_k}^{(k)})$
        \STATE Define the re-noising factor $n_k = (1-s_{k-1})\left(\down(\hat{\bx}_0, 2^{k-1}) - \up(\down(\hat{\bx}_0, 2^k))\right)$
        \STATE Define the next starting point $\hat{\bx}^{(k-1)} = \up(\hat{\bx}^{(k)}) + n_k$
    \ENDFOR
    \STATE \textbf{Return:} $\bx_{K-1}$
    \end{algorithmic}
\end{algorithm}

% \begin{algorithm}[tb]
%    \caption{Pyramidal Flow Sampling \chen{fix this. make $s_k=e_{k+1}$}}
%    \label{pf-sample-code}
%     \begin{algorithmic}[1]
%     \STATE \textbf{Input:} Trained flow model $\bv$, number of stages $K$, list of $K$ start times $s_k$, number of timesteps per stage $N$.
%     \STATE \textbf{Output:} Generated image.
%     \STATE Sample the starting point $\bx_0 \sim \cN\left(0, I \cdot \frac{1}{2^{K-1}}\right)$
%     \FOR{stage $k = 0$ to $K-1$}
%         \FOR{timestep $t=s_k$ to $s_{k+1}$, step = $\frac1N$}
%             \STATE Update the image as $\bx^{(k)} = \bx^{(k)} + \bv(\bx^{(k)}, t) \cdot \frac1N$
%         \ENDFOR
%         \STATE Define the re-noising factor $\alpha = \frac{1-s_k}{\sqrt{2^{K-1-k}}}$
%         \STATE Sample block noise $n \sim \cN(0, \Sigma')$
%         \STATE Define the next starting point $\bx^{(k+1)} = \up(\bx^{(k)}) + \alpha n$
%     \ENDFOR
%     \STATE \textbf{Return:} $\bx_{K-1}$
%     \end{algorithmic}
% \end{algorithm}

\begin{algorithm}[tb]
   \caption{EdifyImage Flow Matching Training}
    \begin{algorithmic}[1]
    \STATE \textbf{Input:} Dataset of images $D$, number of epochs $M$, critical time points $T_1$ and $T_2$, where $0 < T_2 < T_1 < 1$
    \STATE \textbf{Output:} Trained flow model $\bv_t^{(0)}, \bv_t^{(1)}, \bv_t^{(2)}$
    \STATE Initialize the model weights
    \FOR{epoch = 1 to $M$}
        \FOR{$\bx_1$ in $D$}
            \STATE Sample the stage $k$ from a uniform distribution over $\{0, 1, 2\}$
            \STATE Sample $t$ from a uniform distribution over $\left[T_{k+1}, 1\right]$ where $T_0=1$ and $T_3=0$
            \STATE Sample the start point as random Gaussian $\bx_0^{(0)} \sim \cN(0, I^{(0)})$
            \STATE Compute noisy image $\bx_t^{(k)} = (1-t)\down(\bx_0^{(0)}, 2^k) + \mu^{(k)}(\bx_1, t)$ 
            \STATE Compute velocity target $\bu_t^{(k)}(\bx_t^{(k)}|\bx_1)=-\down(\bx_0^{(0)}, 2^{k}) + \frac{d\mu^{(k)}(\bx_1, t)}{dt}$
            \STATE Forward the model to get $\bv_t^{(k)}(\bx_t^{(k)})$
            \STATE Calculate loss $\operatorname{MSE}\left(\bv_t^{(k)}(\bx_t^{(k)}), \bu_t^{(k)}(\bx_t^{(k)}|\bx_1)\right)$
            \STATE Run back-propagation, update network parameters
        \ENDFOR
    \ENDFOR
    \STATE \textbf{Return:} Trained flow model $\bv_t^{(2)}, \bv_t^{(1)}, \bv_t^{(0)}$
    \end{algorithmic}
   \label{alg-EdifyImage-Train}
\end{algorithm}

\begin{algorithm}[tb]
\caption{EdifyImage Flow Matching Sampling}
\begin{algorithmic}[1]
\STATE \textbf{Input:} Trained flow model $\bv_t^{(0)}, \bv_t^{(1)}, \bv_t^{(2)}$, standard Gaussian noise of the largest scale $\bx_0^{(0)}$, 
critical time points $T_1$ and $T_2$
\STATE \textbf{Output:} Largest resolution sample $\hat{\bx}_{1}^{(0)}$ 
\STATE $\hat{\bx}_{T_2}^{(2)} = \operatorname{ODEINT}(\bv_t^{(2)}, \{0, T_2\}, \down(\bx_0^{(0)},4))$ 
\STATE $\hat{\bx}_{T_2}^{(1)} = \up(\hat{\bx}_{T_2}^{(2)}) + (1-T_2)(\down(\bx_0^{(0)}, 2) - \up(\down(\bx_0^{(0)}, 4)))$ \# This is the re-noise step 
%\chen{noise correct?}
% \zelin{revised}
\STATE $\hat{\bx}_{T_1}^{(1)} = \operatorname{ODEINT}(\bv_t^{(1)}, \{T_2, T_1\}, \hat{\bx}_{T_2}^{(1)})$
\STATE $\hat{\bx}_{T_1}^{(0)} = \up(\hat{\bx}_{T_1}^{(1)}) + (1-T_1)(\bx_0^{(0)} - \up(\down(\bx_0^{(0)}, 2))$ 
\STATE $\hat{\bx}_{1}^{(0)} = \operatorname{ODEINT}(\bv_t^{(0)}, \{T_1, 1\}, \hat{\bx}_{T_1}^{(0)})$
\STATE \textbf{Return:} $\hat{\bx}_{1}^{(0)}$
\end{algorithmic}
\label{alg-EdifyImage-Sampling}
\end{algorithm}

\subsection{\textcolor{black}{Discussions with PixelFlow~\citep{pixelflow}}}
\textcolor{black}{Here we discuss the relationship between the concurrent work PixelFlow~\citep{pixelflow} and PyramidalFlow~\citep{PyramidFlow}. The algorithm of PyramidalFlow is shown in~\Cref{pf-train-code}. Line 10 and Line 11 are exactly the same as Equation (2-3) of the PixelFlow paper, while Line 12 is the same as Equation (4) of PixelFlow. PixelFlow also uses a re-noising procedure as in PyramidalFlow, although details are not presented in their paper.}

\subsection{Sampling in Pyramid Flow Matching}
Sampling from a multiscale Pyramidal Flow model is sequential, starting from stage $K-1$ and ending at stage $0$, where $K$ is the number of scales used in training. For each stage, two timesteps are defined: the start time $s_k$ and the end time $e_k$. In this section, we assume the stages to have zero time overlap, that is, $s_{k-1}=e_k$. Also, the stages together span the whole time interval, that is, $s_{K-1}=0$ and $e_0=1$. The noise is sampled once in the largest dimension $n \sim \cN(0, I)$. Within each stage the sampling process follows the standard flow path between $\bx_s=s_{k-1} \cdot \up(\down(\bx, 2^k)) + (1-s_{k-1}) \cdot \down(n, 2^{k-1})$, and $\bx_e=e_{k-1} \cdot \down(\bx, 2^{k-1}) + (1-e_{k-1}) \cdot \down(n, 2^{k-1}).$

The jump between consecutive stages needs to be handled carefully to enforce the consistency of the flow. Consider the jump from stage $k$ to stage $k-1$. The procedure starts with up-sampling the endpoint of stage $k$, yielding $$\up(\bx_e) = s_{k-1}\up(\down(x, 2^k)) + (1-s_k)\up(\down(n, 2^k))$$

The distribution of this data differs from the distribution of the startpoint $\bx_s^{k-1}$ of the $k-1$ stage.

$\down(x, 2^{k-1})\sim \cN\left(0, I \cdot \frac{1}{2^{k-1}}\right)$, and similarly $\down(x, 2^{k})\sim \cN\left(0, I \cdot \frac{1}{2^{k}}\right)$, however the upsampled noise $\up(\down(n, 2^k)) \sim \cN\left(0, \Sigma\cdot \frac{1}{2^{k}}\right)$ where $\Sigma$ is a block-diagonal matrix that has $2\times2$ blocks of ones on the diagonal and zeroes elsewhere. Then if we define a different block-diagonal matrix $\Sigma'$ with blocks

$$
\Sigma'_{block} = \begin{pmatrix}
1&-1\\
-1&1\\
\end{pmatrix}
$$

and set $\alpha = \sqrt{\frac{3}{4^k}}$, and sample $m \sim \cN(0, \Sigma')$, we have $Var(\up(\bx_e^{k}) + (1-s_{k-1})\alpha\cdot m) = Var(\bx_s^{k-1})$. In practice this means that at the end of stage $k$ we sample $m\sim \cN(0, \Sigma')$ and set $x_s^{k-1}=\up(\bx_e^{k}) + (1-s_{k-1})\alpha\cdot m$ as the starting point of stage $k-1$.

\section{More Training Details}
\label{sec-training-details}
\begin{table}[t]
\centering
\caption{\textcolor{black}{\textbf{Training details across datasets and resolutions.} We summarize hyperparameters, computation complexity, and resource usage across benchmarks. Training time and speed may fluctuate due to hardware conditions.}}
\vspace{1mm}
\resizebox{\textwidth}{!}{%
\begin{tabular}{@{}l|c|c|c|c@{}}
\toprule
\textbf{Training Details} & \textbf{CelebA-HQ 256} & \textbf{CelebA-HQ 512} & \textbf{CelebA-HQ 1024} & \textbf{ImageNet 256} \\ 
\midrule
Backbone & DiT-L/2 & DiT-L/2 & DiT-L/2 & DiT-B/2 / DiT-XL/2 \\
Largest latent size & 32 $\times$ 32 & 64 $\times$ 64 & 128 $\times$ 128 & 32 $\times$ 32 \\
Batch size & 256 & 256 & 128 & 256 \\
Init LR & 2e-4 & 2e-4 & 2e-4 & 1e-4 \\
LR schedule & CosineAnnealingLR & CosineAnnealingLR & CosineAnnealingLR & N/A \\
Final LR & 1e-6 & 1e-6 & 1e-6 & N/A \\
Optimizer & AdamW & AdamW & AdamW & AdamW \\
Training epochs & 500 & 1000 & 2000 & 1400/120 \\
Time segments & 0.5 & 0.33, 0.67 & 0.33, 0.67 & 0.5 \\
Number of scales & 2 & 3 & 3 & 2 \\
VAE type & EQVAE & SDVAE & SDVAE & EQVAE \\
EMA decay rate & N/A & N/A & N/A & 0.9999 \\
Hardware & 8$\times$H200 & 8$\times$H200 & 8$\times$H200 & 8$\times$H200 \\
\textcolor{black}{Peak memory per GPU (GB)} & \textcolor{black}{42.2} & \textcolor{black}{82.8} & \textcolor{black}{120.5} & \textcolor{black}{24.9 / 68.3} \\
\textcolor{black}{Training wall-clock (days)} & \textcolor{black}{1.2} & \textcolor{black}{3.1} & \textcolor{black}{8.3} & \textcolor{black}{8.5 / 9.4} \\
\bottomrule
\end{tabular}%
}
\label{tab:training-details}
\end{table}

We present the detailed training settings in~\Cref{tab:training-details}, highlighting key implementation differences across datasets and resolutions. For CelebA-HQ, we utilize a DiT-L/2 backbone across all resolutions while adjusting the number of scales based on complexity—two scales for $256\times256$ and three scales for higher resolutions. For ImageNet, we experiment with both DiT-B/2 and DiT-XL/2 backbones to demonstrate scaling properties. Notably, we use EQVAE for our lower-resolution models due to its superior reconstruction quality, while higher-resolution CelebA-HQ models leverage SDVAE for better stability. All models were trained on 8×H200 GPUs, with batch sizes adjusted according to memory constraints.

\section{Limitations}
\label{sec-limitations}

First, we didn't train our XL model on ImageNet for $7M$ steps due to a restriction of computational resources. This limitation may have prevented our model from reaching its full potential in comparison to diffusion models that are typically trained for significantly longer periods. Second, our approach still requires storing part of the model weights for each scale, which increases memory requirements during training, although our MoT design partially mitigates this issue. Third, while our method shows excellent performance on unconditional and conditional generation tasks, we have not yet extended it to conditional generation scenarios such as text-to-image synthesis or inpainting, which are important applications for generative models. Finally, our method inherits limitations from the VAE architecture used for latent space projection, potentially limiting the fidelity of fine details in generated images. Future work could address these limitations by exploring more efficient parameter-sharing techniques across scales, developing conditional variants of our approach, implementing memory-efficient sampling strategies, and investigating improved latent space representations.

\section{Impact Statement}
\label{sec-impact-statement}
This paper presents Laplacian Multi-scale Flow Matching (\modelname), a novel approach for generative image modeling. We believe this work has several potential positive impacts on the research community and society:

Our method improves computational efficiency in high-resolution image generation, potentially reducing the energy consumption and carbon footprint associated with training and deploying generative models. By requiring fewer function evaluations and reduced GFLOPs, \modelname contributes to more sustainable AI development.

The technical innovations in multi-scale representations and unified transformer architectures may inspire new approaches to other generative tasks beyond images, such as audio, video, and 3D content creation. The proposed mixture-of-transformers architecture with parameter sharing could influence more efficient model designs across various domains.

However, we acknowledge that advances in generative image modeling also come with potential societal concerns. Like other generative models, this technology could be misused to create misleading or deceptive content. We encourage responsible development and deployment of systems based on our research, including content safeguards and transparent disclosure of AI-generated media.

Our work focuses on advancing the technical capabilities of generative models rather than specific applications. We emphasize that any practical implementation should consider ethical implications and be developed in accordance with responsible AI principles.
\end{document}